\documentclass[manuscript,screen]{acmart}
\AtBeginDocument{%
  }

\usepackage{graphicx} 
\usepackage{amsmath,amsfonts} 
\usepackage{algorithm} 
\usepackage{algpseudocode}
\usepackage{textcomp} 
\usepackage{xcolor} 
\usepackage{soul} 
\usepackage{balance} 
\usepackage{hyperref} 
\usepackage{orcidlink} 
\usepackage[mathlines,switch]{lineno} 
\usepackage{url} 
\usepackage{mathtools} 
\usepackage{multirow} 
\usepackage{pgfplots} 
\usepackage{tabularx} 
\usepackage{booktabs} 
\usepackage{caption} 
\usepackage{subcaption} 
\usepackage{footnote} 
\usepackage{float} 
\usepackage{adjustbox} 
\usepackage{colortbl}
\usepackage[listings,skins,breakable]{tcolorbox}
\usepackage{tikz}
\usetikzlibrary{positioning, calc, shapes.geometric}
\usepackage[framemethod=default]{mdframed}
\usepackage{xcolor}

\setcopyright{acmlicensed}
\copyrightyear{2026}
\acmYear{2026}
\acmDOI{XXXXXXX.XXXXXXX}

\newboolean{showcomments}
\setboolean{showcomments}{true}
\ifthenelse{\boolean{showcomments}}
{
	\definecolor{myyellow}{RGB}{255, 228, 26}
	\definecolor{myblue}{RGB}{50, 50, 220}
	\newcommand{\nb}[2]{
		{\sf
			\fcolorbox{myyellow}{yellow}{\scriptsize\textbf{#1}}%
			$\blacktriangleright$%
			{\color{myblue}\fontsize{7pt}{8pt}\selectfont\textbf{#2}}%
		}%
	}
}
{
	\newcommand{\nb}[2]{}
}

\begin{document}

\title{FuzzInsight: Navigating the Efficiency-Effectiveness Trade-off in Coverage-Guided DNN Fuzzing}

\author{Vivek Vekariya}
\email{vivek.vekariya@tum.de}
\orcid{0009-0005-1482-7142}
\affiliation{%
  \institution{Technical University of Munich}
  \city{Munich}
  \state{Bayern}
  \country{Germany}
}
\author{Fathiyeh Faghih}
\email{fathiyeh.faghih@bcu.ac.uk}
\orcid{0000-0002-8877-6895}
\affiliation{%
  \institution{Birmingham City University}
  \city{Birmingham}
  \state{West Midlands}
  \country{United Kingdom}
}
\author{Alexander Pretschner}
\email{alexander.pretschner@tum.de, pretschner@fortiss.org}
\orcid{0000-0002-5573-1201}
\affiliation{%
  \institution{Technical University of Munich \& fortiss GmbH}
  \city{Munich}
  \state{Bayern}
  \country{Germany}
}

\renewcommand{\shortauthors}{Vekariya et al.}

\begin{abstract}
Collecting and labeling diverse test datasets for testing the reliability of Deep Neural Networks (DNNs) is both time-consuming and expensive. This challenge has driven significant interest in coverage-guided fuzzing techniques for DNNs, which effectively generate defect-revealing corner cases from existing limited test datasets. The efficiency and effectiveness of these campaigns heavily depend on configuration choices, including coverage criteria and seed-selection strategies.

We present a configuration-oriented analysis and decision guidelines for scalable DNN fuzzing that focuses on two key bottlenecks: the underutilization of available input seeds during fuzzing due to seed-selection strategies, and the identification of computationally efficient coverage criteria that effectively guide the discovery of high-value corner cases. Our findings indicate that existing seed selection strategies repeatedly favor a small subset of available seeds, leaving a substantial fraction of collected inputs unexplored. To address this, we propose Equalized Sampling, a constraint-based strategy that achieves a mean seed queue utilization of 77\%, improving upon prior approaches by 13 percentage points (p.p.). 

We further evaluate the cost-effectiveness of coverage guidance across 84 distinct fuzzing configurations on four well-established computer vision benchmark datasets: MNIST, SVHN, GTSRB, and CIFAR-100. In our evaluated settings, non-structural coverage criteria, LSCD, LSC, and MDSC, achieve coverage outcomes comparable to both structural and other non-structural criteria, while incurring substantially lower computational cost. Moreover, these criteria enable broad behavioral exploration of the DNN under test, reducing the need to explicitly monitor multiple coverage metrics. Among these, LSCD exhibits a consistent but modest runtime advantage, while LSCD, LSC, and MDSC collectively require substantially lower runtime over other criteria.

Finally, we demonstrate that retraining with the generated corner cases yields substantial average accuracy gains in generalization robustness up to 77 p.p. without harming model stability on the original test distribution. Together, these results provide a scalable blueprint for practitioners to optimize DNN testing strategies by balancing computational efficiency with rigorous model hardening.
\end{abstract}


\maketitle

\section{Introduction}

To assess the quality of Deep Neural Networks (DNNs), test engineers evaluate their ability to generalize to unseen scenarios using metrics such as accuracy, precision, recall, etc., on a test dataset. Trusting the performance of DNNs solely based on these metrics can be misleading, primarily due to concerns about the quality of the test dataset ~\cite{testing_ml_industry_icse21, structured_verification_ml_rahul, ml_testing_survey, 2020-Riccio-EMSE}. A high-quality test dataset should closely capture and reflect real-world in-field properties for the trustworthy assessment of DNNs. However, collecting and labeling test data requires significant time and resources ~\cite{testing_ml_industry_icse21, structured_verification_ml_rahul, ml_testing_survey, test_optimization_survey, 2020-Riccio-EMSE}. 

Like traditional software systems, DNNs can exhibit undesired or erroneous behavior due to hidden, unidentified defects that often manifest in corner cases: rare inputs residing in low-probability regions of the input space. In this work, we focus on corner cases discovered via fuzzing. We start with correctly classified inputs from the test set and apply realistic perturbations from the operational domain (e.g., subtle changes in lighting, rotation, or sensor-related noise). A generated input is considered a corner case if it successfully triggers a misclassification, thereby exposing a region of the input space where the DNN lacks robustness. An important factor influencing test dataset quality is the extent to which it includes corner cases, which are essential for exposing such defects and limitations in DNNs’ ability to generalize. Including such corner case data can enhance test data quality and give test engineers an opportunity to address and improve DNN robustness. 

Prior surveys and industrial reports suggest growing interest in automated corner case generation for DNNs using Coverage-Guided Fuzzing (CGF) ~\cite{structured_verification_ml_rahul, ml_testing_survey, testing_ml_industry_icse21, 2020-Riccio-EMSE}. Inspired by traditional software fuzzing, CGF has effectively alleviated the challenges of test data generation and corner-case identification for DNNs \cite{deepgauge, tensorfuzz, deephunter, deeptest, 2025-Chen-EMSE}. However, the data‑centric nature of DNNs and the broad spectrum of available coverage criteria make it difficult to determine an efficient and effective fuzzing configuration in practice. Most existing DNN fuzzing frameworks, such as DeepHunter ~\cite{deephunter} or TensorFuzz ~\cite{tensorfuzz}, primarily rely on structural coverage criteria. These white-box approaches calculate coverage based on internal neuron activations, requiring full access to the DNN’s architecture. In contrast, emerging non-structural criteria such as Surprise Coverage (SC)~\cite{surprise_adquacy, surprise_adequacy_journal, refinement_sa, revisit_coverage} and Latent Space Class Dispersion (LSCD)~\cite{lscd_intro} work in a gray-box manner and do not require complete access to the DNN architecture. Specifically, gray-box criteria rely on high-level representations (e.g., logits or latent vectors) to summarize DNN behavior without monitoring individual neuron activations. This can make them more scalable and easier to apply in practice. Despite their demonstrated effectiveness in evaluating dataset quality under various fault conditions ~\cite{taxonomy_faults, lscd_intro}, these non-structural metrics remain largely unexplored for guiding test data generation in fuzzing.

Practitioners face an under-specified configuration problem when applying coverage-guided fuzzing to DNNs: selecting coverage criteria and seed-selection strategies under practical resource constraints and the permissible level of model access. Existing DNN fuzzing research provides limited empirical insight into how these factors jointly affect fuzzing effectiveness and scalability. Accordingly, this paper contributes a configuration-oriented empirical analysis and a practitioner-oriented decision guideline for scalable DNN fuzzing. We derive this guidance by empirically comparing and incorporating non‑structural coverage criteria (such as surprise coverage and latent space class dispersion) alongside established structural metrics. Beyond simple corner-case generation, our empirical study aims to characterize efficiency-effectiveness trade-offs across configurations for DNN testing by investigating three key aspects of fuzzing campaign:    

\begin{itemize}
\item \textbf{Seed Queue Utilization:} Many prior DNN fuzzing evaluations and default configurations initialize the fuzzing campaigns with a randomly selected subset of the available test data as the initial seeds ~\cite{deepgauge, tensorfuzz, deephunter, deeptest, DLFuzz}. While functional, this approach can leave large portions of the available test data unselected for fuzzing, including inputs that might expose distinct failure modes. Also, simply increasing the initial seed count does not resolve this inefficiency. The true bottleneck lies in the fuzzer's internal seed-selection strategy. If it repeatedly prioritizes a small subset of seeds, many others receive little or no mutation budget, resulting in "missed opportunities" to discover corner cases from those seeds.

Prior work rarely reports how many distinct initial seeds actually contribute to corner case discovery after a fuzzing campaign, nor how the seed-selection strategy affects this contribution ~\cite{tensorfuzz, deephunter}. This motivates our study of seed queue utilization, a metric defined as the fraction of distinct initial seeds that contribute to corner-case discovery. To fully assess the exploration potential of each strategy, we evaluate utilization using the full available labeled test dataset. A low seed queue utilization indicates that the mutation budget concentrates on a small subset of seeds, leaving many available seeds effectively unexercised during fuzzing. We first conduct a critical evaluation of existing seed selection strategies to quantify this inefficiency ~\cite{tensorfuzz, deephunter}. We address the identified gaps by introducing \emph{Equalized Sampling}, a simple constraint-based seed-selection strategy. From a practitioner’s perspective, seed queue utilization is not merely a configuration choice or an end goal itself, but a useful diagnostic of whether the mutation budget is allocated to only a subset of seeds during fuzzing instead of leveraging the potential of all available input seeds.

\item \textbf{Efficiency and Cross-Coverage Synergies:} A major barrier to scaling DNN fuzzing is the high computational overhead of monitoring multiple, complex coverage criteria simultaneously. This creates an \emph{efficiency-effectiveness trade-off}: computationally intensive criteria may provide more informative guidance but slow down fuzzing campaigns, whereas lightweight criteria improve scalability but may yield weaker feedback. Our goal is to identify whether specific coverage metrics can indirectly achieve broader test thoroughness. To this end, we investigate \emph{cross-coverage}, a phenomenon in which the targeted maximization of one coverage metric during fuzzing tends to increase other coverage criteria. This approach acknowledges the interconnected nature of neural and coverage behaviors: when a fuzzer is guided to explore behaviors captured by one metric, it may also stimulate behaviors reflected by other criteria.

Our analysis reveals that non-structural metrics offer a dual advantage. First, as gray-box measures, they do not require full access to the DNN's internal architecture and offer computational efficiency in generating test cases significantly faster than structural alternatives. Second, they exhibit meaningful cross-coverage synergies by increasing several other coverage criteria without directly optimizing them. Identifying such coverage alignment can help practitioners support broader DNN behavioral exploration without having to monitor multiple complex metrics.

\item \textbf{Model Robustness and Practical Utility:}  
To demonstrate that our fuzzing framework provides actionable value beyond optimizing configuration choices, we evaluate the utility of the generated corner cases for enhancing DNN robustness. We close the testing loop by retraining the model on the generated corner-case data and measuring the resulting impact through two key properties. We evaluate \emph{Model Stability}, defined as the preservation of accuracy on the original test set, and \emph{Generalization Robustness}, defined as the accuracy gain on held-out fuzzed data that the model has not previously encountered. This evaluation aims to demonstrate that the fuzzing process yields a measurably more robust model, bridging the gap between automated test generation and tangible model improvement. 
\end{itemize}
Our empirical study spans 7 coverage criteria and 3 seed selection strategies across four widely recognized computer vision benchmark datasets: MNIST \cite{mnist}, SVHN \cite{svhn}, GTSRB \cite{gtsrb}, and CIFAR-100 ~\cite{cifar100}. Our evaluation demonstrates that existing seed selection strategies often struggle to capitalize on the diversity of the available seed pool. Our proposed Equalized Sampling strategy improves seed queue utilization by balancing the allocation of the mutation budget across the available seed pool. It achieves the highest seed queue utilization among all evaluated methods. On average, it utilizes 77.58\% of the available seeds for corner case discovery across all datasets, compared with 64.61\% for the widely used DeepHunter strategy and just 6.99\% for TensorFuzz. This represents gains of 12.97 and 70.59 percentage points (p.p.) over DeepHunter and TensorFuzz, respectively. Higher utilization means a larger fraction of initial seeds actively participate in the fuzzing process, receive mutation budget, and contribute to the discovery of corner cases.

Among the studied criteria, the three gray-box latent-space metrics, Latent Space Class Dispersion, Likelihood-based Surprise Coverage (LSC), and Mahalanobis Distance-based Surprise Coverage (MDSC), consistently exhibit the lowest computational overhead, particularly as model complexity increases. Across all studied datasets, LSCD completes fuzzing in approximately 128 minutes, compared with 148 minutes for LSC and 152 minutes for MDSC. This makes LSCD approximately 1.16 times (i.e., 16\%) faster than LSC, 1.19 times (i.e., 19\%) faster than MDSC, and 8.71 times faster than Neuron Coverage (NC), the fastest structural alternative. In contrast, k-Multisection Neuron Coverage (k-MNC) is the slowest coverage criterion, requiring an average of 2 days, 4 hours, and 29 minutes to complete fuzzing across all datasets; LSCD is therefore 24.61 times faster than k-MNC. Importantly, these efficiency gains do not come at the expense of coverage effectiveness. We observe large differences in computational cost but comparatively smaller differences in the evaluated coverage outcomes, highlighting the efficiency-effectiveness trade-off. Our cross-coverage analysis shows that lightweight latent-space criteria achieve coverage values comparable to those attained when structural and non-structural criteria are used directly as guidance. This enables practitioners to achieve broad test thoroughness without having to monitor multiple complex metrics.

Our evaluation further reveals that all studied coverage criteria effectively generate corner cases that harden the DNN via retraining. They yield broadly comparable improvements in generalization robustness, with a mean accuracy gain of 77.53 p.p. and specific accuracy gains across benchmarks ranging from 61.51 to 86.78 p.p. Importantly, this enhancement is achieved without harming the original model's stability. The mean accuracy drop on the original test set is only 0.37 p.p., with a maximum drop of 2.35 p.p. relative to the baseline DNNs. This marginal change indicates that the retraining process avoids overfitting to the newly generated corner cases, ensuring that the DNN under test retains its original stability while simultaneously gaining broader generalization robustness. Differences in robustness gains across coverage criteria are modest, indicating that no single metric consistently dominates in effectiveness.

Taken together, these results show that although coverage effectiveness and robustness gains differ only modestly across the evaluated criteria, their computational overheads vary substantially. In our evaluated settings, LSCD, MDSC, and LSC offer a favorable efficiency-effectiveness trade-off, achieving coverage and robustness outcomes comparable to those of other criteria at significantly lower computational costs. Among these three, LSCD exhibits a consistent but moderate runtime advantage. This distinction is practically relevant because the criteria differ not only in the time required to complete a fuzzing campaign but also in the amount and type of additional data they need to maintain during fuzzing: LSCD relies primarily on class centroids, MDSC additionally requires a covariance matrix, and LSC depends on kernel density estimation over latent representations. As a result, these criteria offer practical, resource-aware options for large-scale DNN fuzzing, enabling effective model hardening without the substantial overhead of traditional structural criteria.

These findings motivate a resource-aware configuration guidance for DNN fuzzing that improves the utility of available data while reducing computational cost. In summary, this paper makes the following contributions:

\begin{itemize}
\item We provide practitioner-oriented guidelines for navigating the efficiency-effectiveness trade-offs in DNN fuzzing under model-access and compute constraints. The guidelines are grounded in an empirical study of 21 distinct fuzzing configurations on four established vision benchmarks (MNIST, SVHN, GTSRB, and CIFAR-100), yielding 84 distinct fuzzing setups. 

\item We investigate non-structural coverage criteria (such as surprise-coverage variants and LSCD) for guidance in fuzzing. Our results show that non-structural metrics achieve comparable cross-coverage gains, while LSCD, MDSC, and LSC consistently incur the lowest computational overhead.

\item We evaluate existing seed selection strategies and introduce Equalized Sampling, which improves seed queue utilization by enabling a larger portion of the available seeds to actively contribute to corner‑case discovery.

\item We demonstrate the practical utility of the generated corner cases by retraining DNNs on fuzzed data, showing substantial gains in generalization robustness on unseen fuzz splits while preserving model stability on the original datasets.

\item We provide our framework, "FuzzInsight," alongside all experimental data and generated corner cases in our replication package to promote community research ~\cite{replication-package}.
\end{itemize}
\section{Coverage-Guided Fuzzing in DNNs}\label{sec:approach}
\begin{figure*}[htbp]
  \centering
  \includegraphics[width=\linewidth, keepaspectratio]{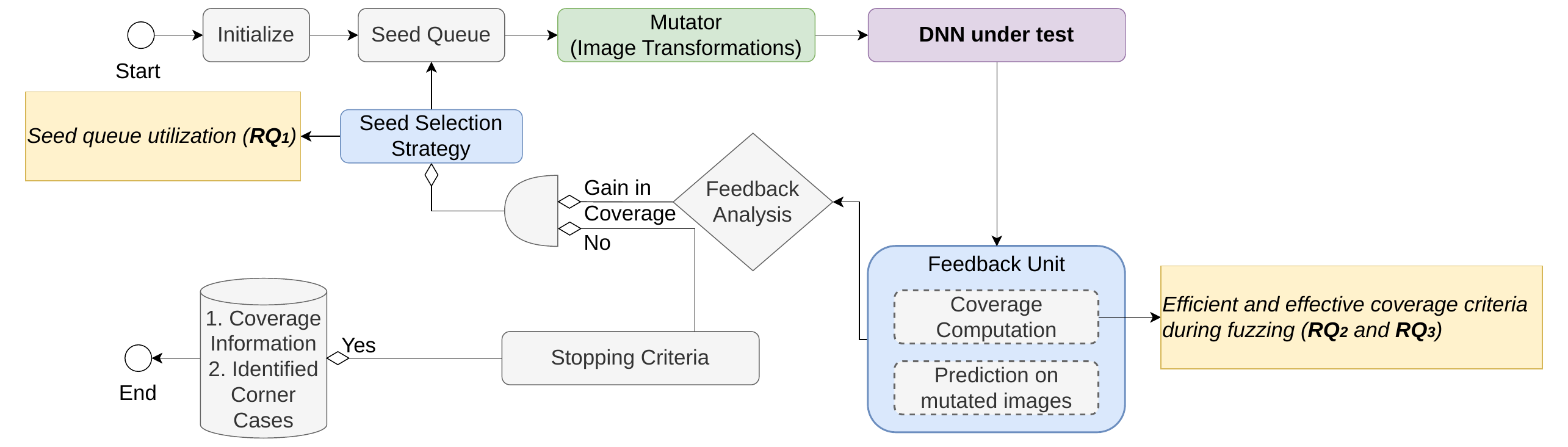} 
  \caption{Overview of coverage-guided fuzzing for DNNs.}
  \label{fig:fuzzing}
\end{figure*}

Inspired by established software fuzzing methods, fuzzing for testing DNNs aims at systematically finding the erroneous behaviors of the DNN under test. In software testing, fuzzing relies on various coverage criteria to ensure comprehensive code execution. In a similar way, coverage criteria for DNNs are successful in the systematic investigation of defects and identification of corner cases in a DNN under test \cite{deephunter, deepxplore, deepgauge}. Various industry surveys highlight that practitioners use CGF techniques such as DeepHunter \cite{deephunter} to identify corner cases and uncover defects in a DNN under test \cite{structured_verification_ml_rahul, ml_testing_survey, testing_ml_industry_icse21, 2020-Riccio-EMSE}. By generating high-quality corner cases from an existing seed pool, CGF significantly reduces the substantial costs associated with manually collecting and labeling new test data. This approach effectively leverages existing data to enhance the quality of the test dataset for comprehensive DNN testing. Our study utilizes a modified fuzzing framework based on DeepHunter as illustrated in \autoref{fig:fuzzing}.

\textbf{Initialization:} 
A fuzzing campaign begins with selecting "initial seeds" from available test data. Our approach selects test inputs for which the DNN exhibits nominal behavior (i.e., its output matches the ground truth). This choice ensures that the fuzzer operates on inputs for which the DNN behaves as expected, allowing mutations to reveal deviations from this behavior. Inputs that already exhibit abnormal behavior prior to fuzzing are naturally identifiable as failure cases and are therefore not useful for fuzzing. In contrast to previous works ~\cite{deephunter, tensorfuzz} that use a randomly selected subset of the test dataset, our work uses the entire available test dataset. As we will discuss in \autoref{sec:empirical-study}, this holistic approach provides a rigorous baseline for evaluating how different seed selection strategies during fuzzing maximize the utility of available test inputs.

\textbf{Seed Queue:}
The selected initial seeds form a seed queue, which is dynamically managed using various seed selection strategies. These strategies play a critical role in determining which specific test inputs from the queue will be chosen for subsequent processing by the mutators, which then apply various transformations.

\textbf{Mutator:} 
Selected seeds are then systematically mutated by applying various transformations aiming to explore the internal structure of the DNN using the described coverage criteria. In our study, the mutator applies domain-specific transformations such as brightness, contrast, fog, rain, motion blur, zoom blur, etc., widely used to test the functional robustness of DNNs in safety-critical applications such as Automated Driving (AD) ~\cite{hendrycks_framework, image_perturbations_stocco, 2022-Stocco-TSE}. These transformations are intended to approximate environmental and sensor-related corruptions, such as changes in lighting conditions, atmospheric effects, and motion-induced degradation, which are relevant for safety-critical perception tasks such as automated driving. 

A common challenge in automated fuzzing is the risk of generating transformed inputs with a high degree of perturbation ~\cite{2020-Stocco-ICSE, 2023-Riccio-ICSE}. Previous approaches often mitigate this by applying strict $L_p$ norm constraints (e.g., $L_0$ or $L_\infty$) to discard inputs that deviate significantly from the original seed. While such constraints are useful for controlling perturbation magnitude, they are closely aligned with adversarial-style input generation and may restrict broader image-level corruptions such as illumination changes, weather effects, motion blur, or sensor noise.
In contrast, our goal is to evaluate the robustness of DNNs under domain-relevant transformations rather than adversarial robustness under strict $L_p$ constraints. Although these transformations may exceed strict $L_p$-norm bounds, they capture image-level variations commonly considered in computer vision robustness testing, such as illumination changes, weather effects, motion blur, and sensor noise. We therefore use commonly adopted robustness transformations that affect the image more globally, allowing the fuzzer to explore input variations aligned with environmental changes studied in robustness evaluation~\cite{hendrycks_framework, image_perturbations_stocco, 2022-Stocco-TSE}.

\textbf{Coverage Computation:} 
Fuzzing requires a DNN-specific model profile to compute various coverage criteria. It is a dictionary containing the profile of every trainable neuron in the DNN, based on the training data. The profile contains values such as mean, squared mean, standard deviation, and lower and upper activation bounds that are used for the computation of structural coverage criteria. For non-structural criteria, it includes kernel density estimators, covariance matrices, class centroids, etc.

Current state-of-the-art fuzzing frameworks utilize various structural coverage criteria, such as neuron coverage, k-Multisection NC, and Neuron Boundary Coverage (NBC), to guide the fuzzer ~\cite{deepxplore, deepgauge, deephunter}. A neuron is considered covered by NC if its activation value exceeds the defined threshold $t$. The k-MNC criterion divides the activation range $[\mathit{low, high]}$ of each trainable neuron derived from the training set into $k$ equal sections. A section is considered covered if the neuron is activated in that range using the generated test input. For NBC, the range $[\mathit{-\infty, low]}$ and $[\mathit{high, +\infty]}$ is divided into $N$ sections ~\cite{deepgauge}.  

In our research, we take a step further by including non-structural criteria such as surprise coverage and latent space class dispersion during fuzzing ~\cite{surprise_adequacy_journal, lscd_intro}. These criteria are already popular and widely used for test dataset quality assessments ~\cite{refinement_sa, surprise_adequacy_journal, lscd_intro}. They offer a different perspective by quantifying the "surprise" of a given test input based on the latent space vectors. By incorporating these, we aim to empirically evaluate their effectiveness in guiding the generation of diverse and challenging test cases using fuzzing.

Likelihood-based surprise coverage is computed in two primary steps using \autoref{lsc1} and \autoref{lsc2}. First, a density function $\hat{f}(\cdot)$ is estimated via Kernel Density Estimation (KDE) over the latent space vectors $\overrightarrow{\alpha_n}(T_{train})$ extracted from the $n^{th}$ layer of the DNN for the training set $T_{train}$ (\autoref{lsc1}). $K_H$ denotes the Gaussian kernel and $x_i$ represents individual training samples of class $i$. For a given test input $x$, the surprise score is then derived as the negative logarithm of this estimated density (\autoref{lsc2}). This transformation ensures that inputs from low-density regions are those furthest from the training distribution, yielding higher LSC. The cumulative LSC for the entire test set $X$ is then calculated following \autoref{dsa6}.

\begin{equation}\label{lsc1}
\hat{f}(x) = \frac{1}{|\overrightarrow{\alpha_N}(T_{train})|} \sum_{x_i \in T_{train}} K_H(\overrightarrow{\alpha_N}(x) - \overrightarrow{\alpha_N}(x_i))
\end{equation}

\begin{equation}\label{lsc2}
\mathit{LSC}(x) = -\log(\hat{f}(x))
\end{equation}

Distance-based Surprise Coverage (DSC) for a given input $x$ is calculated based on the Euclidean distance of its latent space vectors from its nearest neighbor within the same predicted class $c_x$ (\autoref{dsa2}) and from other output classes (\autoref{dsa4}). The nearest neighbours $x_a$ and $x_b$ for $x$ are calculated using \autoref{dsa1} and \autoref{dsa3} respectively. In the given equations, $D$ is the DNN under test, $C$ is the total number of output classes, $D(.)$ is the prediction by DNN for a given input, and  $\overrightarrow{\alpha_N}(.)$ represents latent space vectors corresponding to a given input.

\begin{equation}\label{dsa1}
x_a = \arg\min_{D(x_i) = c_x} \|  \overrightarrow{\alpha_N}(x) - \overrightarrow{\alpha_N}(x_i) \|
\end{equation}

\begin{equation}\label{dsa2}
\text{dist}_a = \| \overrightarrow{\alpha_N}(x) - \overrightarrow{\alpha_N}(x_a) \|
\end{equation}

\begin{equation}\label{dsa3}
x_b = \arg\min_{D(x_i) \in C \setminus \{c_x\}} \| \overrightarrow{\alpha_N}(x_a) - \overrightarrow{\alpha_N}(x_i) \|
\end{equation}

\begin{equation}\label{dsa4}
\text{dist}_b = \| \overrightarrow{\alpha_N}(x_a) - \overrightarrow{\alpha_N}(x_b) \|
\end{equation}

\begin{equation}\label{dsa5}
\text{DSC($x$)} = \frac{\text{dist}_a}{\text{dist}_b}
\end{equation}

Mahalanobis Distance-based Surprise Coverage extends distance-based surprise coverage by incorporating the covariance matrix derived using latent space vectors of training data for each class. Instead of measuring deviation using Euclidean distance alone, MDSC computes how far a test input lies from the distribution of its predicted class $c_x$ using the Mahalanobis distance using \autoref{mdsc1}. Let $\mu_i$ denote the mean latent representation of class $i$ computed from the training data, and let $\Sigma_i$ denote the corresponding covariance matrix. For a given test input $x$ belonging to class $c_x$, the MDSC score is calculated as follows:

\begin{equation}\label{mdsc1}
\mathit{MDSC}(x) = 
\left(\overrightarrow{\alpha_N}(x) - \mu_{c_x}\right)^T 
\Sigma^{-1} 
\left(\overrightarrow{\alpha_N}(x) - \mu_{c_x}\right)
\end{equation}

For the entire test dataset $X$, LSC/DSC/MDSC is calculated using \autoref{dsa6}. The calculated coverage value segments are divided into  \( k \) segments. These segments are defined by an upper bound $U$ and buckets \( B = \{b_1, b_2, \ldots, b_k\} \) that divide the interval \( (0, U] \).

\begin{equation}\label{dsa6}
\text{LSC/DSC/MDSC($X$)} = \frac{\left| \left\{ b_i \mid \exists x \in X : LSC/DSC/MDSC(x) \in \left( U \cdot \frac{i-1}{k}, U \cdot \frac{i}{k} \right] \right\} \right|}{k} * 100
\end{equation}

LSCD characterizes latent-space behavior using a simpler formulation based on class centers $\overrightarrow{c_i}$, which represent the mean latent vectors of training inputs for each class. Unlike DSC, which relies on pairwise distances between samples, and MDSC, which additionally normalizes distances along high class covariance direction using the covariance matrix, LSCD measures the average dispersion of samples around their respective class centers.

For a given class $i$, the value $LSCD_i$ in \autoref{eq1} captures the average Euclidean distance of test inputs belonging to that class from the corresponding class center. Here, $\overrightarrow{s_i}^{(j)}$ denotes the latent representation of the $j^{th}$ input of class $i$ and $dist$ is Euclidean distance. The overall LSCD value is obtained by averaging these class-wise dispersion values across all output classes $C$.

\begin{equation}\label{eq1}
	\begin{multlined}
		\mathit{LSCD}_i = \frac{{\sum\limits_{j=0}^{|s_i|} \text{$dist$}(\overrightarrow{c_i}, \overrightarrow{s_i}^{(j)})}}{{|s_i|}}; 		\ \ \ \ \ \ \ \mathit{LSCD}= \frac{\sum\limits_{i=0}^{C} \mathit{LSCD}_i}{C}
\\
	\end{multlined}
\end{equation}

Compared to DSC and MDSC, LSCD requires only class centroids computed from training data, avoiding pairwise computations and covariance estimation. This could make LSCD computationally lightweight and easier to scale in diverse setups. Higher LSCD values indicate that test inputs are increasingly dispersed from the training distribution of their associated class. The increased class dispersion in the latent space during testing indicates that the DNN encounters inputs whose features are less familiar or not clearly identified than those seen during training. This is vital for reliable testing of DNNs, as real-world inputs will inevitably exhibit feature variations that may not be similar to the training data. Using diverse transformations during mutation increases coverage across different criteria. This allows the fuzzer to exercise a broader range of DNN behaviors, which is important for identifying diverse defects and corner-case behavior in the model under test.

\textbf{Feedback Analysis:} 
The mutated inputs that increase coverage are kept in the seed queue; otherwise, they are discarded. The increase in structural coverage refers to the activation of new neuron behaviors (e.g., previously uncovered sections of a neuron's activation range under k-MNC or NBC). For surprise-based coverage, a gain in coverage is observed if the mutated image yields a higher LSC/DSC/MDSC value in \autoref{lsc2}, \autoref{dsa5}, and \autoref{mdsc1} than its parent seed. The increase in SC score indicates that the latent space vector of the mutated image is located farther from its nearest neighbor in the current output class and from other classes, in the low-variance dimension of the training distribution or in a lower likelihood region of the KDE. This positioning often places the mutant image closer to decision boundaries, indicating a more "surprising" input. For LSCD, a gain in coverage is observed when the latent space vector of a mutated image has a higher Euclidean distance compared to its parent seed from its ground-truth class. This implies the mutant image's latent space vectors are becoming more dispersed, moving further away from the training clusters without variance-based normalization, unlike MDSC. 

In our study, if the mutant image results in a false classification, it is flagged as a corner case and is taken out of the seed queue. This feedback loop, combining coverage gains with prediction on mutant images, allows the fuzzing process to strategically generate corner case data that exposes diverse behaviors of the DNN under test.

\textbf{Seed Selection Strategy:}
Based on the feedback analysis, the seed selection strategies determine which seed will go next for the mutation. In this paper, we scrutinize two prominent strategies presented in their chronological order of introduction, and subsequently propose a new hybrid strategy. 

TensorFuzz, developed by Odena et al., pioneered the concept of CGF for DNNs ~\cite{tensorfuzz}. It is based on the seed selection strategy described in ~\autoref{tensorfuzz} that uses a dynamic "reservoir" $R$ of size $m$ (default $m=6$) ~\cite{tensorfuzz}. In each iteration, $R$ consists of $m-1$ seeds selected from the rear of the queue $Q$ of size $n$ to prioritize recently mutated seeds, augmented by one seed $s_{\text{random}}$ sampled uniformly from the entire queue $Q$. The next seed $s_{\text{next}}$ is then selected from this reservoir using a uniform sampling function $U$.

\begin{equation}\label{tensorfuzz}
	\begin{multlined}
R = {s_{n-m+2}, \dots, s_n} \cup {s_{\text{random}}}; \ \ \ \quad s_{\text{random}} \sim U(Q); \ \ \ s_{\text{next}} \sim U(R)
	\end{multlined}
\end{equation}

The DeepHunter framework assigns a probability of selecting a seed $P(.)$ to each mutated seed $s$ ~\cite{deephunter}. $P(s)$ is determined based on how many times a seed has been mutated using  \autoref{deephunter_prob} ~\cite{deephunter}. $g(s)$ represents the number of times the given seed is mutated. $\gamma$ is a normalization factor that constrains the decrease ratio of the probability, and $p_{\text{min}}$ is the minimum probability that a seed can be selected. Initially, $p_{\text{min}}$ is set to be high for new seeds to ensure that they are considered 'interesting' enough to be included in a queue and improve coverage. This probability is later updated and adjusted based on how frequently the seed is mutated. This selection promotes diversity in selecting input seeds during fuzzing.

\begin{equation} \label{deephunter_prob}
P(s) =
\begin{cases}
    1 - g(s)/\gamma, & \text{if } g(s) < (1 - p_{\text{min}}) \times \gamma \\
    p_{\text{min}}, & \text{otherwise}
\end{cases}
\end{equation}

\textbf{\textit{Note:}} Both of the above-mentioned strategies differ in how they utilize mutation information during seed selection. TensorFuzz does not account for the mutation frequency of individual parent seeds, as it relies primarily on recency through its reservoir-based sampling. On the other hand, DeepHunter uses mutation frequency at the level of individual mutated seeds by reducing the probability of frequently selected seeds. However, this mechanism operates locally at the level of individual seeds and may not ensure a balanced, global distribution of mutation budget across all initial seeds.

This oversight can lead to a form of "selection bias" at the level of parent seeds. Seeds that consistently contribute to increased coverage are more likely to be selected repeatedly, while others receive limited or no mutation opportunities. As a result, the fuzzing process may disproportionately focus on a subset of initial seeds, ignoring others that could potentially uncover different types of faults in the DNN under test. This is particularly problematic in practice, where initial seeds are often curated by domain experts or selected by automated methods to represent high-value, diverse scenarios ~\cite{seed_selection, deephyperion, test_selection_methods, distribution_aware_test_selection}. Although all initial seeds are used for baseline testing of a DNN, seed selection strategies that fail to utilize them adequately during fuzzing underutilize their generative potential. In this context, the human and computational effort invested in acquiring a diverse dataset is under-leveraged for the specific task of discovering new corner cases.

Our working assumption is that each input seed represents a distinct sample from the data distribution and should therefore have a fair opportunity to contribute to fuzzing. Ensuring that seeds receive mutation budget more evenly helps prevent the fuzzer from over-exploiting only a small subset of seeds while ignoring others that may expose different failure modes. To the best of our knowledge, the impact of the seed selection strategies during fuzzing has rarely been quantified, as previous studies were typically limited to randomly selected subsets ~\cite{deephunter, deepxplore, deeptest}.

To address these gaps, we examine the seed queue utilization under various seed selection strategies using the full available test dataset. We hypothesize that evaluating a fuzzer against the full dataset is essential to establishing a performance ceiling, a baseline representing the fuzzer’s maximum potential for uncovering unique corner cases. By quantifying the gap between the available seeds and those actually selected for mutation, we can identify exactly where current seed selection strategies fail to leverage the provided input seeds. As discussed before, the existing strategies often neglect the mutation frequency of individual parent seeds, leading to a feedback loop in which coverage contributing seeds may be oversampled while the rest of the queue remains effectively unexplored.

To mitigate this selection bias, we propose a constraint-based sampling strategy called "Equalized Sampling", detailed in ~\autoref{alg:fuzz_eq_sampling}. The primary objective of our approach is to promote a more balanced distribution of the mutation budget across all initial seeds. We implement this by introducing a selection constraint based on the mutation frequency of each parent seed, thereby giving underutilized seeds a fairer chance of being selected and mutated. Before selecting a new seed, the $mutated\_count$ for the selected seed is checked (how often its parent seed is mutated) (Line 5 in ~\autoref{alg:fuzz_eq_sampling}). If it exceeds the maximum times ($max\_times$), the next seed from the queue is selected to avoid bias (Line 10 in ~\autoref{alg:fuzz_eq_sampling}). We set the default value to $max\_times$ = 100, based on our preliminary sensitivity analysis. To evaluate the impact of this parameter, we also conducted a sensitivity analysis across a range of values [50, 250] (detailed in ~\autoref{sec:results}). Within this observed window, equalized sampling demonstrated high stability, with seed-queue utilization varying by only $\pm$1\% across all studied datasets. While practitioners may adjust this threshold based on the dataset's scale or the intended number of fuzzing iterations, our results suggest that sampling efficiency remains robust across configurations. Higher values allow for more extensive mutations per seed, whereas lower values accelerate the rotation through the seed queue. In this way, Equalized Sampling aims to enhance the overall seed queue utilization during the fuzzing process, especially in large and diverse datasets.

\begin{algorithm}
\caption{Equalized Sampling Algorithm}
\label{alg:fuzz_eq_sampling}
\small
\begin{algorithmic}[1]
\Function{equalized\_sampling}{queue, mutation\_frequency\_dict, max\_times}
    \While{True}
        \State \texttt{cur\_seed} $\gets$ \texttt{queue.get()}
        \State \texttt{mutated\_count} $\gets$ \texttt{mutation\_frequency\_dict[cur\_seed]}
        \If{\texttt{mutated\_count} $\le$ \texttt{max\_times}}
            \State \texttt{mutation\_frequency\_dict[cur\_seed]} $\gets$ \texttt{mutated\_count + 1}
            \State Update \texttt{cur\_seed.probability}
            \State \Return \texttt{cur\_seed}
        \Else
            \State \texttt{cur\_seed} $\gets$ \texttt{queue.next()}
        \EndIf
    \EndWhile
\EndFunction
\end{algorithmic}
\end{algorithm}

\textbf{Stopping Criterion:} 
The fuzzer runs for a defined maximum number of iterations or until the seed queue is empty. We run the fuzzer for a maximum of 15,000 iterations for our experiments. After running a fuzzing campaign for 15,000 iterations, we found that the coverage criteria used for each fuzzing configuration reached a saturation level. Continuing with the same criteria might lead to diminishing returns in fuzzing efficiency. This is because the same criteria will likely identify fewer and fewer corner cases over time.

\section{Empirical Study} \label{sec:empirical-study}
As discussed in the introduction, coverage‑guided DNN fuzzing inherently involves trade‑offs between data utility, computational cost, and model hardening. Building on this motivation, our empirical study systematically examines how concrete configuration choices influence these trade‑offs in practice. To this end, our investigation aims to answer the following research questions.

\noindent
\textbf{RQ\textsubscript{1} (Seed Queue Utilization):} 
\textit{How do different seed selection strategies affect the utilization of available input seeds during fuzzing?}

Our first research question, RQ\textsubscript{1}, aims to identify which seed selection strategy effectively utilizes a more diverse set of available seeds during fuzzing. Although all seeds are accessible during a fuzzing campaign, selection strategies determine which seeds participate in mutation. When a strategy disproportionately prioritizes a small subset of available seeds, much of the seed pool's diversity remains underexplored, limiting the discovery of corner cases, particularly in large, complex datasets.

\noindent
\textbf{RQ\textsubscript{2} (Efficient coverage criteria for  fuzzing):} 
\textit{Which coverage criterion is the most efficient within coverage-guided fuzzing for DNNs?}

The corner case identification using CGF is directly linked to the coverage criteria used. The complexity of these criteria directly influences the computational time required for a fuzzing campaign. Thus, our second research question, RQ\textsubscript{2}, aims to find the most computationally efficient coverage criteria for fuzzing in DNNs.

\noindent
\textbf{RQ\textsubscript{3} (Effectiveness in cross-coverage maximization):}
\textit{How do different coverage criteria compare in terms of self-coverage and cross-coverage when used to guide DNN fuzzing?}

Beyond computational efficiency, an effective coverage criterion should enable broader and more comprehensive testing. Specifically, we investigate whether guiding fuzzing with certain coverage criteria consistently triggers high cross-coverage (i.e., indirectly increasing other metrics without directly optimizing them), while also achieving high self-coverage (i.e., maximizing the selected coverage criterion itself). If a specific guidance criterion reliably yields high coverage values across diverse structural and non-structural metrics, it indicates strong cross-coverage gains. Therefore, RQ\textsubscript{3} explicitly evaluates the effectiveness of various coverage criteria w.r.t. achieving both high self and cross-coverage.

\noindent
\textbf{RQ\textsubscript{4} (Generalization \& Stability):} What is the quantitative impact of retraining with fuzzed corner cases on the DNN's generalization robustness versus original stability?

We also aim to measure the practical utility of the data generated using fuzzing in RQ\textsubscript{4}. A successful fuzzing campaign must demonstrate a significant robustness gain (accuracy improvement on fuzzed data) while maintaining a near-zero drop in the DNN's original performance (accuracy on the original test set).

\subsection{Datasets and Deep Neural Networks} \label{sec:dnn_architecture}
We conducted our empirical study on widely studied computer vision benchmarks and image classification DNNs, selected to span different levels of complexity and domain characteristics, thereby enhancing the reliability and generalizability of our findings. We start with the MNIST dataset ~\cite{mnist}. It is a collection of handwritten digits. Along with the widely studied LeNet-5~\cite{lenet_5} architecture, this setup provides a well‑established baseline for evaluating fuzzing techniques. To increase the complexity of DNNs and datasets under more challenging conditions, we employ the SVHN ~\cite{svhn} and GTSRB ~\cite{gtsrb} datasets. SVHN (Street View House Numbers) introduces real-world complexity by capturing diverse house number sequences from Google Street View. GTSRB (German Traffic Sign Recognition Benchmark) features traffic signs captured under various illumination shifts, rotations, and occlusions characteristic of AD. It has 43 distinct output classes corresponding to one distinct traffic sign, making it ideal for evaluating DNNs in autonomous driving contexts. For these datasets, we utilize high-performance architectures based on Spatial Transformer Networks (STNs), which are specifically designed to handle the geometric and spatial variance inherent in these perception tasks. Finally, we also incorporate the CIFAR-100 ~\cite{cifar100} dataset using the VGG-16 architecture ~\cite{vgg16}. It consists of 100 distinct output classes, including various animals, vehicles, and household objects, providing significantly higher intra-class variance and label granularity than other benchmarks. In our study, CIFAR-100 serves as a practical proxy for complex, large-scale benchmarks. By monitoring the DNN under test across all 100 output classes, we ensure our findings on fuzzing effectiveness are based on a comprehensive assessment of broad test thoroughness. This approach demonstrates how our framework maintains its discovery potential when navigating high-dimensional spaces, providing a clear path for studying scalability trends in higher-dimensional classification settings.

\begin{table}[h]
\caption{Overview of Datasets and DNN Performance}
\label{tbl:dataset_dnn_overview}
\centering
\Large 
\begin{tabular}{lcccccc}
\toprule
\textbf{Dataset} & \multicolumn{3}{c}{\textbf{Number of Images}} & \textbf{Output} & \textbf{Acc.} \\
\cmidrule(r){2-4}
& \textbf{Train} & \textbf{Val} & \textbf{Test} & \textbf{Classes} & \textbf{(in \%)} \\
\midrule
MNIST     & 48,000 & 12,000 & 10,000 & 10  & 98.98 \\
SVHN      & 62,268 & 10,989 & 26,032 & 10  & 92.83 \\
GTSRB     & 35,610 & 3,870  & 12,631 & 43  & 99.69 \\
CIFAR-100 & 40,000 & 10,000 & 10,000 & 100 & 74.00 \\
\bottomrule
\end{tabular}
\end{table}

\begin{figure}[ht]
\centering
\begin{tikzpicture}[
    node distance=1.0cm,
    block/.style={rectangle, draw, fill=blue!10, text width=2.5cm, text centered, rounded corners, minimum height=0.75cm},
    stn_block/.style={rectangle, draw, fill=orange!10, text width=2.5cm, text centered, rounded corners, minimum height=0.75cm},
    vgg_block/.style={rectangle, draw, fill=green!10, text width=2.6cm, text centered, rounded corners, minimum height=0.75cm},
    arrow/.style={thick, ->, >=stealth}
]

\node [block] (lenet_in) {\textbf{LeNet-5} \\ (MNIST)};
\node [block, below=0.45cm of lenet_in] (lenet_conv) {2$\times$ Conv + AvgPool};
\node [block, below=0.45cm of lenet_conv] (lenet_fcblock) {FC Block};
\node [block, below=0.45cm of lenet_fcblock] (lenet_lastfc) {Final FC (10)};
\node [block, below=0.45cm of lenet_lastfc] (lenet_softmax) {Softmax};

\draw [arrow] (lenet_in) -- (lenet_conv);
\draw [arrow] (lenet_conv) -- (lenet_fcblock);
\draw [arrow] (lenet_fcblock) -- (lenet_lastfc);
\draw [arrow] (lenet_lastfc) -- (lenet_softmax);

\node [stn_block, right=2.5cm of lenet_in] (stn_in) {\textbf{STN} \\ (SVHN/GTSRB)};
\node [stn_block, below left=0.8cm and -1.0cm of stn_in] (loc) {Localization Network};
\node [stn_block, below right=0.8cm and -1.0cm of stn_in] (main_flow) {3$\times$ Conv Block};
\node [stn_block, below=2.3cm of stn_in] (stn_fcblock) {FC Block};
\node [stn_block, below=0.45cm of stn_fcblock] (stn_lastfc) {Final FC (10/43)};
\node [stn_block, below=0.45cm of stn_lastfc] (stn_softmax) {Softmax};

\draw [arrow] (stn_in) -- (loc);
\draw [arrow] (stn_in) -- (main_flow);
\draw [arrow] (loc) |- (stn_fcblock);
\draw [arrow] (main_flow) -- (stn_fcblock);
\draw [arrow] (stn_fcblock) -- (stn_lastfc);
\draw [arrow] (stn_lastfc) -- (stn_softmax);

\node [vgg_block, right=2.5cm of stn_in] (vgg_in) {\textbf{VGG-16} \\ (CIFAR-100)};
\node [vgg_block, below=0.45cm of vgg_in] (vgg_b1) {Block 1--2 \\ (4$\times$ Conv + BN)};
\node [vgg_block, below=0.45cm of vgg_b1] (vgg_b2) {Block 3--5 \\ (9$\times$ Conv + BN)};
\node [vgg_block, below=0.45cm of vgg_b2] (vgg_fcblock) {FC Block};
\node [vgg_block, below=0.45cm of vgg_fcblock] (vgg_lastfc) {Final FC (100)};
\node [vgg_block, below=0.45cm of vgg_lastfc] (vgg_softmax) {Softmax};

\draw [arrow] (vgg_in) -- (vgg_b1);
\draw [arrow] (vgg_b1) -- (vgg_b2);
\draw [arrow] (vgg_b2) -- (vgg_fcblock);
\draw [arrow] (vgg_fcblock) -- (vgg_lastfc);
\draw [arrow] (vgg_lastfc) -- (vgg_softmax);

\node[above=0.2cm of lenet_in] {\textit{Small-scale}};
\node[above=0.2cm of stn_in] {\textit{Mid-scale (Spatial)}};
\node[above=0.2cm of vgg_in] {\textit{Large-scale}};

\end{tikzpicture}
\caption{Architectural overview of the evaluated DNNs, ranging from classic convolutional stacks (LeNet-5) to spatial transformer networks and modular VGG-style networks. The final FC layer size corresponds to the number of output classes for each dataset: 10 for MNIST/SVHN, 43 for GTSRB, and 100 for CIFAR-100. Final FC layer is used to compute non-structural coverage criteria.}
\label{fig:dnn_architectures}
\end{figure}
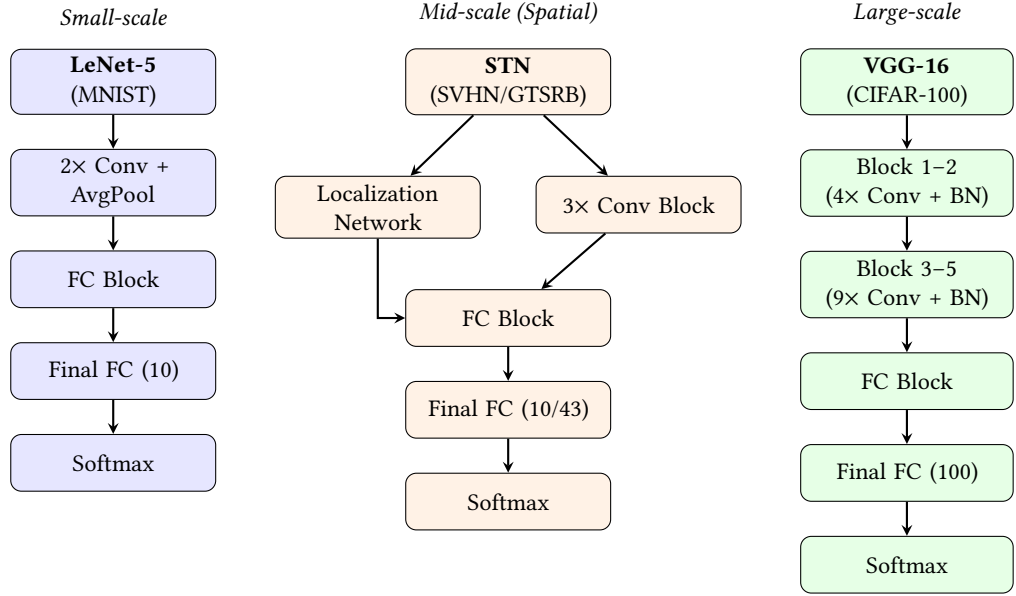

The DNNs used for all four datasets are trained using the Adam~\cite{adam} optimizer with a learning rate scheduler for 50 epochs (MNIST), 100 epochs (SVHN and GTSRB) and 200 epochs (CIFAR-100). We use an early stopping mechanism to prevent overfitting. The performance of DNNs and the dataset overview are shown in ~\autoref{tbl:dataset_dnn_overview}. The high-level architectural overview for the evaluated DNNs is illustrated in \autoref{fig:dnn_architectures}. These models represent a spectrum of complexity, from the classic convolutional layers of LeNet-5 to the parallel localization branches of the Spatial Transformer Network and the deep, modular blocks of VGG-16. For clarity, we omit layer-by-layer parameters here. However, the complete architectural specifications, including exact layer dimensions, kernel sizes, and pre-trained weights, are provided in our replication package~\cite{replication-package}.

\subsection{Evaluation Procedure} \label{sec:procedure}

For each dataset, we configure 21 distinct fuzzing configurations per dataset that combine 3 seed selection strategies for each of the 7 selected coverage criteria. Therefore, our empirical study analyzes results derived from a total of 84 unique fuzzing configurations. 

We conduct fuzzing experiments using three seed selection strategies: TensorFuzz, DeepHunter (Probability-based), and Equalized Sampling as discussed in \autoref{sec:approach}. We pair each seed selection strategy with NC, k-MNC, NBC, LSC, DSC, MDSC, and LSCD as coverage criteria during fuzzing. The selection of three structural coverage criteria is motivated by their ability to capture a neuron's activation range (from $-\infty$ to $+\infty$) at various granularities and segmentations. At the same time, the choice of non-structural coverage criteria, LSC, DSC, MDSC, and LSCD, is based on their demonstrated effectiveness and widespread use in evaluating test dataset quality under diverse fault conditions. We run each fuzzing experiment for 15,000 iterations. After each successful fuzzing experiment, we first analyze the generated corner cases to determine how many distinct initial seeds were used during fuzzing. To answer RQ\textsubscript{1}, we calculate the seed queue utilization using \autoref{seed_utilization}.

\begin{equation}
\label{seed_utilization}
\text{Seed Queue Utilization (\%)} = \frac{\text{Number of Distinct Initial Seeds}}{\text{Total Initial Seeds}} \times 100
\end{equation}

Based on the design complexity of each coverage criterion, the time required to complete each fuzzing experiment may vary. Thus, we report the total time needed to complete the fuzzing experiment to address RQ\textsubscript{2}. This allows us to identify the most scalable coverage criterion for high-dimensional models. To assess whether a single metric can drive broader testing thoroughness, we conduct a cross-coverage analysis in RQ\textsubscript{3}. Notably, we include Strong Neuron Activation Coverage (SNAC) with other metrics as an independent coverage metric in this analysis. SNAC measures the extent to which inputs push neuron activations into high-activation regions beyond the training baseline. Because SNAC is a functional subset of NBC, using it as a primary guidance criterion in fuzzing would yield only limited functional divergence compared to NBC-guided fuzzing. Instead, by evaluating how other metrics influence SNAC, we can determine their effectiveness for exploring the activation distribution without the overhead of monitoring every individual neuron or meeting coverage criteria. 

\begin{figure*}[htbp]
  \centering
  \includegraphics[width=\linewidth, height=5.1cm, keepaspectratio]{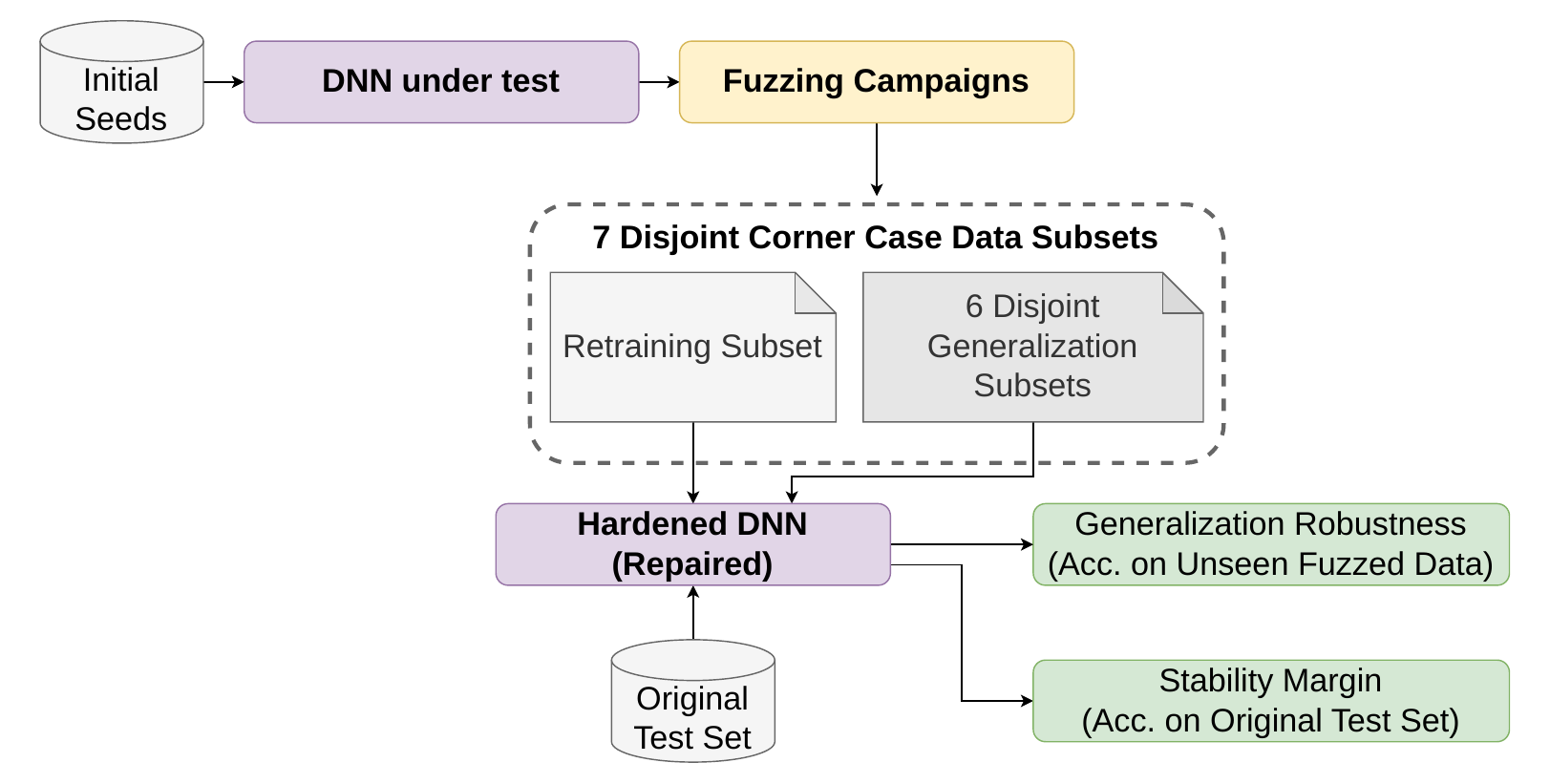} 
  \caption{Overview of re-training pipeline using generated corner cases.}
  \label{fig:retraining_dnn}
\end{figure*}

Finally, in RQ\textsubscript{4}, we evaluate the practical usefulness of the generated corner cases via a model-hardening pipeline as illustrated in ~\autoref{fig:retraining_dnn}. To ensure a rigorous and independent evaluation, we adopt a cross-criterion validation strategy rather than standard $n$-fold cross-validation. While $n$-fold validation assesses stability across random partitions of a single set, our strategy tests the DNN's ability to generalize across entirely disjoint data subsets produced by different coverage criteria during fuzzing. Specifically, the DNN is retrained on corner cases generated by a single coverage metric and subsequently evaluated across five disjoint subsets produced by the remaining criteria. We measure generalization robustness by calculating the accuracy gain on held-out corner case data that the DNN never encountered during retraining. This approach helps assess that the induced robustness generalizes across different objective functions and is not an artifact of the specific manifold explored by a single guidance metric. Following the established methodology ~\cite{data_distillation, suspicious_nn_analysis, structured_verification_ml_rahul}, we utilize the original test set exclusively as a nominal seed pool to evaluate DNN stability. It is measured by evaluating the retrained model on the original, nominal test set. This ensures that the hardening process does not induce catastrophic forgetting or overfitting to the corner case distributions generated during the fuzzing campaign.

\section{Results} \label{sec:results}

\subsection{RQ\textsubscript{1} (Seed Queue Utilization)}

\begin{figure}[t!]
  \centering
  \includegraphics[width=0.72\linewidth, keepaspectratio]{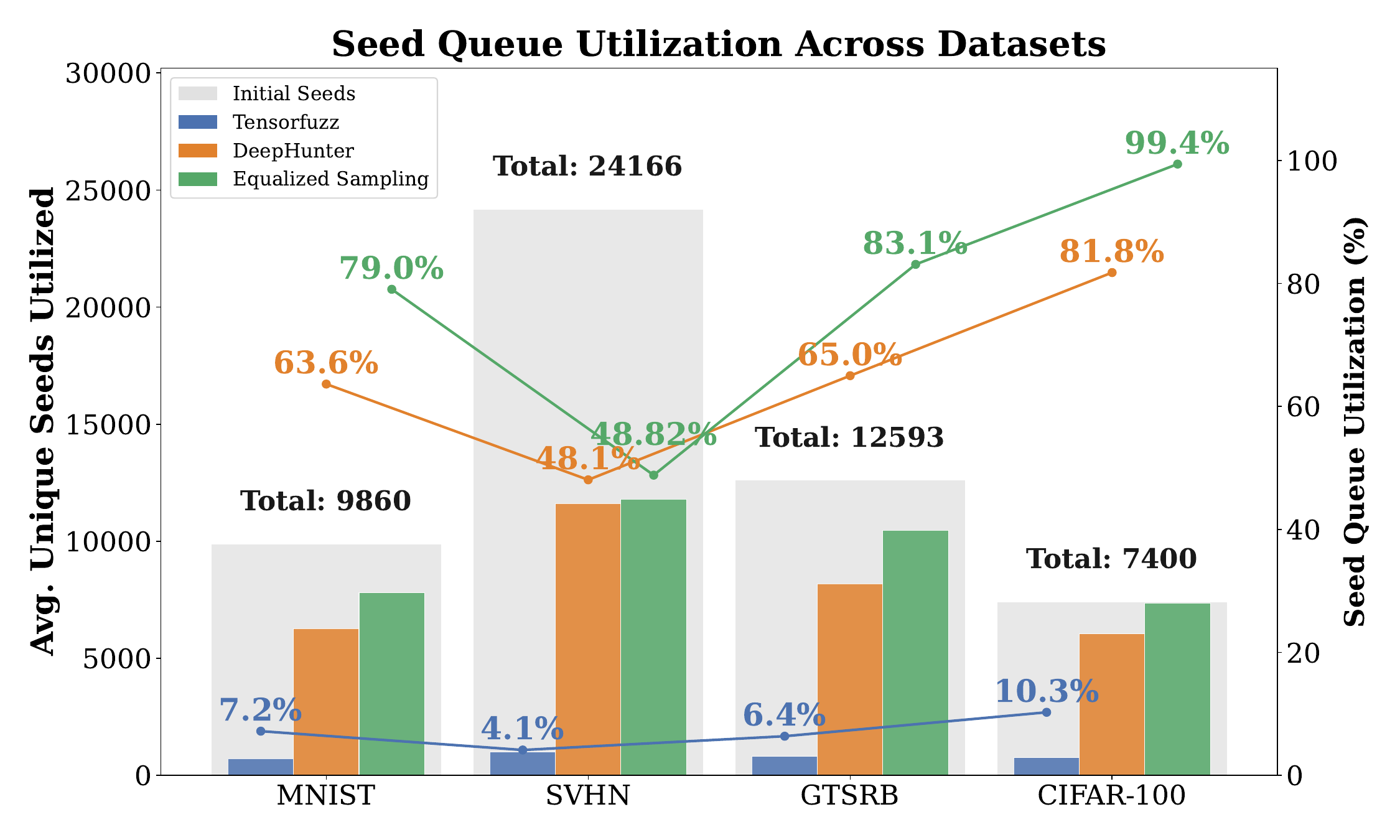}
  \caption{Seed queue utilization across datasets using various seed selection strategies.}
  \label{fig:seed_queue_utilization}
\end{figure}

\autoref{fig:seed_queue_utilization} illustrates the seed queue utilization of each seed selection strategy to capitalize on the available labeled test data. Across the benchmarks, the number of correctly classified inputs available as initial seed is 9,860 for MNIST, 24,166 for SVHN, 12,593 for GTSRB, and 7,400 for CIFAR‑100. By utilizing the full dataset, we ensure a fair comparison across all strategies and avoid the selection bias inherent in smaller subsets. For clarity's sake, we report seed queue utilization as an aggregation across all evaluated coverage criteria, as the variance across criteria was negligible. A comprehensive breakdown of individual criterion-specific utilization is provided in the Appendix for further reference ~\cite{replication-package}.

TensorFuzz exhibits the lowest seed queue utilization across all datasets. This indicates a significant underuse of the available initial seeds during fuzzing. It utilized less than 4.1\% of initial seeds in the SVHN dataset and peaked at only 10.3\% for CIFAR-100. This implies that roughly 90\% of the initial seeds are never selected for mutation during the fuzzing campaign, highlighting a substantial waste of the available labeled data. This is due to its reliance on a reservoir consisting of recently mutated seeds and randomly selected global seeds, which restricts the full exploration of the initial seeds in identifying diverse corner cases.

DeepHunter's probability-based seed selection approach addresses this problem by assigning a probability to mutated seeds based on their mutation frequency. Specifically, in the MNIST dataset, DeepHunter utilized 6,274 of the 9,860 initial seeds, achieving an average utilization of 63.63\% across all coverage criteria. On the SVHN and GTSRB datasets with a large number of initial seed images, it utilizes, on average, 48.10\% and 65.00\% of the initial seeds, respectively. While this approach certainly increases seed queue utilization, we hypothesized a potential limitation: over extended fuzzing periods, this strategy might lead to the seed queue being dominated by frequently mutated seeds. This could eventually reduce the chances for other, less-mutated initial seeds to be explored. Therefore, we introduce a constraint based on the maximum number of times an initial seed can be mutated, ensuring that all initial seeds enter the queue at least once, thereby increasing the seed queue utilization.

As \autoref{fig:seed_queue_utilization} shows, our proposed seed selection strategy - "Equalized Sampling"- using $max\_times = 100$ utilizes the maximum number of initial seeds across all datasets. For instance, it used up to 99.40\% of the initial seeds in the CIFAR-100 dataset, with a minimum of 48.82\% in the SVHN dataset. The introduced constraints in the equalized sampling effectively promote the use of previously underutilized seeds, ensuring they appear in the queue at least once during fuzzing. A distinct trend is observed in the SVHN and GTSRB datasets. In the GTSRB dataset, equalized Sampling maintains a lead at 83.10\% compared to DeepHunter's 65.00\%. However, both strategies exhibit nearly the same performance in the SVHN dataset, with equalized sampling (48.82\%) and DeepHunter (48.10\%). This suggests that, for this specific high-variance dataset, both strategies encounter a similar "bottleneck" in which a large portion of the initial seed pool fails to trigger new coverage, regardless of the selection logic used. Despite this, equalized sampling remains the most robust approach, eliminating 'selection bias' and ensuring the fuzzer explores the entire breadth of the available test suite. 

We will analyze the impact of higher seed selection utilization on DNN hardening later in (\autoref{results:ablation-study}). We first assess the robustness of Equalized Sampling and ensure that the observed gains are not sensitive to a particular hyperparameter choice through a sensitivity analysis of the $max\_times$ parameter. This analysis uses LSCD as the guiding coverage criterion due to its low computational overhead, enabling efficient exploration of the parameter space without altering the seed‑selection behavior under study. As shown in \autoref{fig:ablation_study}, we observed negligible variation ($\pm$1\%) in seed queue utilization using equalized sampling across the tested $[50, 250]$ range. This relative insensitivity suggests that the primary benefit stems from the presence of a selection constraint, which ensures the search during fuzzing eventually rotates through the available seed pool rather than the specific threshold value itself. While the exact choice of $max\_times$ may not be a primary driver of overall utilization, maintaining a finite bound remains important to prevent the fuzzer from repeatedly focusing on a small subset of high‑yield seeds. In contrast, prior strategies lacking such constraints concentrate exploration on these seeds, leading to the low utilization levels reported in \autoref{fig:seed_queue_utilization}. In practice, $max\_times$ may be adjusted based on dataset specificity and the desired search granularity: lower values prioritize faster rotation across the seed pool, while higher values allow for more extensive mutations per seed.

\begin{figure}[htbp]
  \centering
  \includegraphics[width=0.72\linewidth, keepaspectratio]{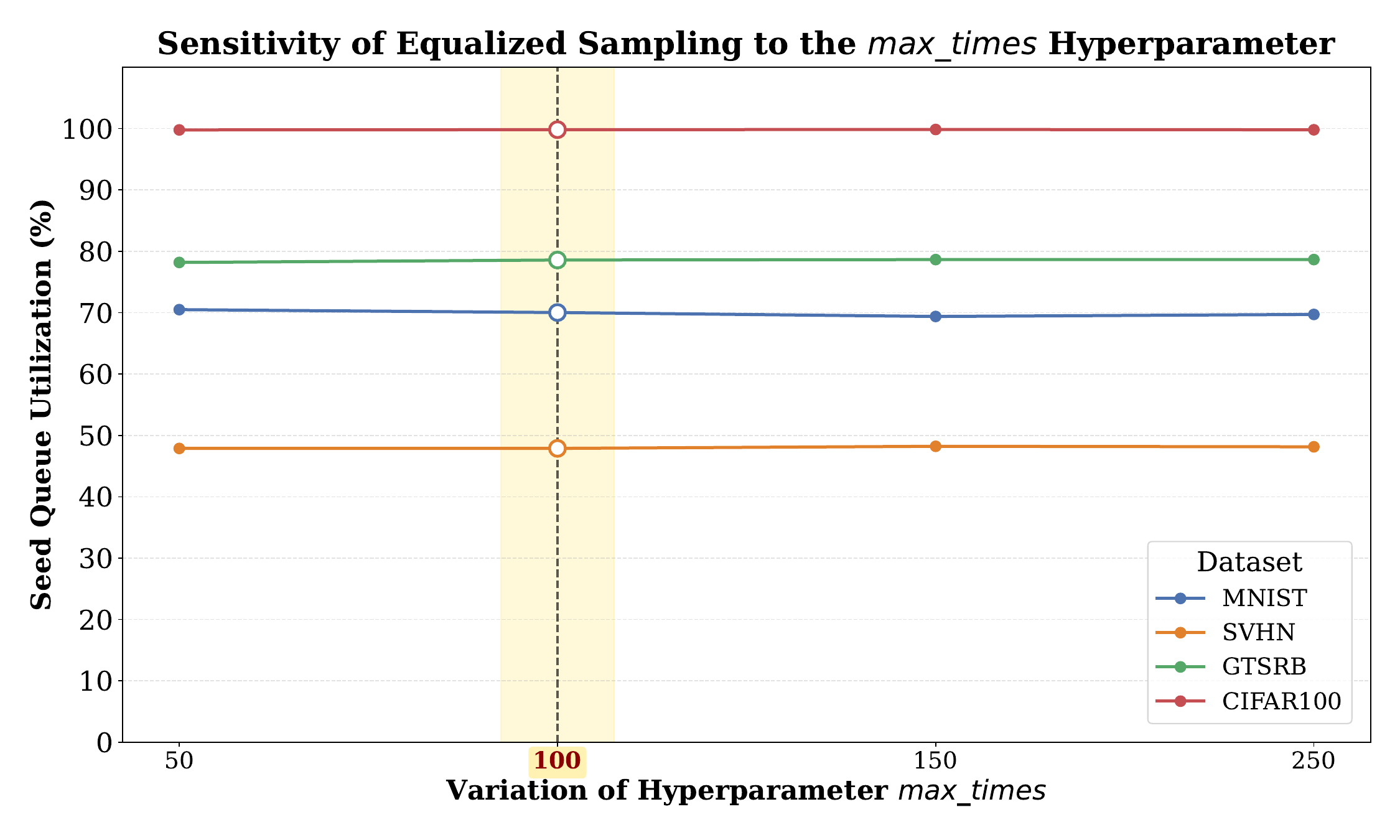}
  \caption{Impact of the ($max\_times$) on equalized sampling strategy.}
  \label{fig:ablation_study}
\end{figure}

\tcbset{breakable = false, enhanced jigsaw, left=1mm, right=1mm, bottom=1mm, boxsep=1mm}
\begin{tcolorbox}[title=\textbf{Answer to RQ\textsubscript{1}: Seed Queue Utilization}]
\textit{Seed selection strategies vary substantially in how effectively they leverage the available seed pool during fuzzing. Equalized Sampling achieves higher seed queue utilization than existing seed selection strategies across all evaluated datasets. It achieves an average seed queue utilization of 77.58\% across all datasets, which is 12.97 p.p. improvement over DeepHunter (64.61\%) and outperforms TensorFuzz (6.99\%).}
\end{tcolorbox}

\subsection{RQ\textsubscript{2} (Efficient coverage criteria for fuzzing)}

\begin{figure}[h]
  \centering
  \includegraphics[width=0.72\linewidth, keepaspectratio]{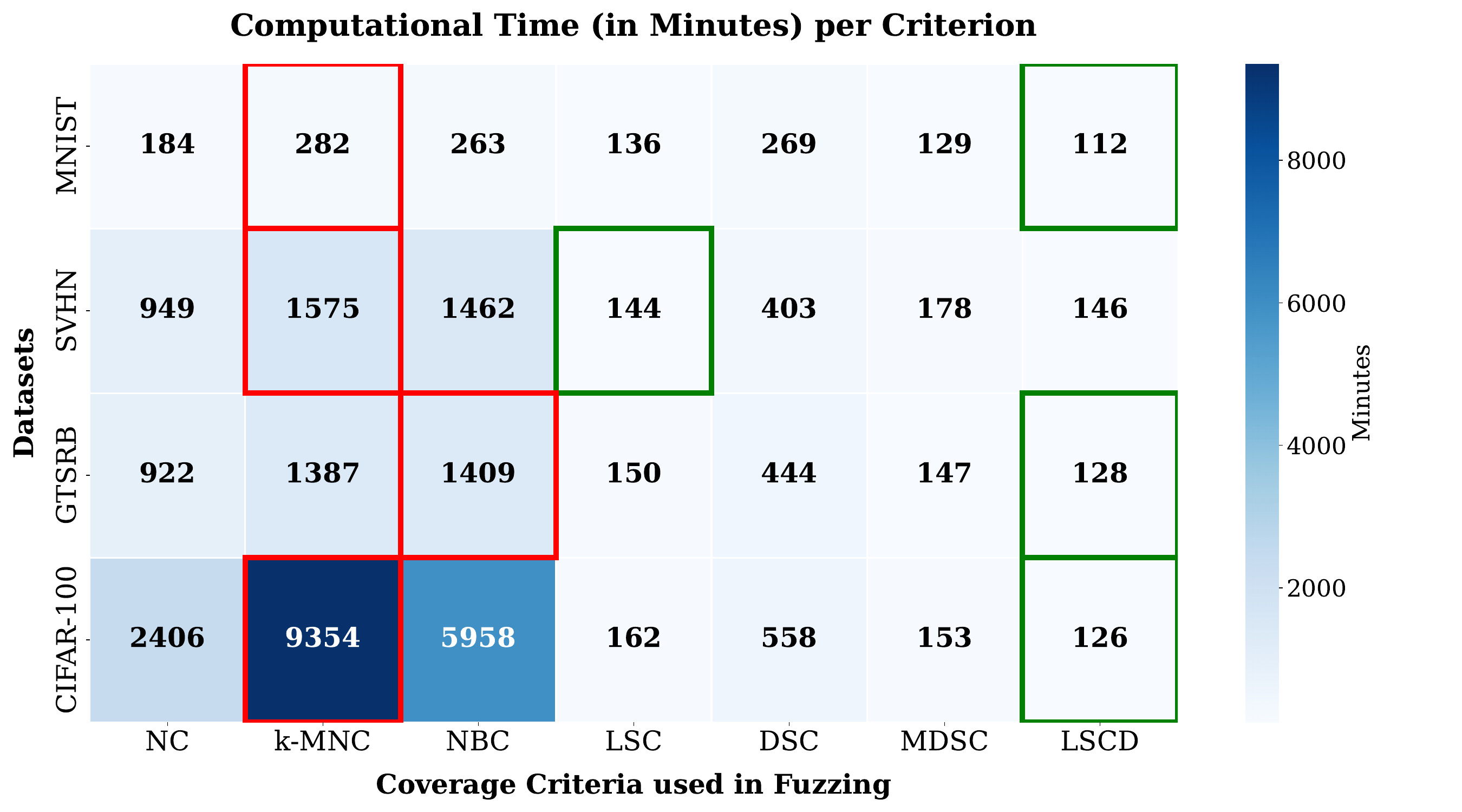}
  \caption{Comparison of computational time required by various coverage criteria during fuzzing.}
  \label{fig:compute_time}
\end{figure}

The design of coverage criteria determines the total time required to complete a fuzzing campaign. In \autoref{fig:compute_time}, the red and green blocks highlight the most time-intensive and fastest criteria per dataset, respectively. All experiments were conducted on an Ubuntu 22.04 system equipped with an AMD Ryzen 2920X 12-core processor, an NVIDIA GeForce RTX 2070 SUPER GPU, and 128 GB of RAM.

The time required to compute structural coverage criteria varies with model complexity. The LeNet-5 DNN architecture used for MNIST image classification is much simpler than the STN and VGG-16 architectures. Consequently, structural coverage computation for LeNet-5 is considerably faster than others, a trend that can be clearly seen in \autoref{fig:compute_time}. Among the structural coverage criteria, NC requires the least computation time as it merely considers whether a neuron's activation surpasses a defined threshold or not. For MNIST, fuzzing with NC as the coverage criterion takes 3 hours 4 minutes (184 minutes) to complete, compared to approximately 1 day 16 hours (2406 minutes) for CIFAR-100. In contrast, k-MNC and NBC demand the most computation time. For the k-MNC and NBC calculations, we used k=1500, which is commonly used in previous work \cite{deephunter}. The higher computation time is due to the need to check whether each of the 1500 sections for each trainable neuron has been covered by previous test inputs. The total time required for fuzzing using k-MNC is 6 days, 11 hours, 54 minutes (9354 minutes) on CIFAR-100 due to the complex architecture like VGG-16 and the design of coverage criteria.

On the other hand, non-structural coverage criteria are consistently more efficient than structural criteria, as they rely primarily on latent-space representations rather than tracking neuron-level activations. This efficiency gap becomes more pronounced as model complexity increases. As indicated by the green highlighted blocks in \autoref{fig:compute_time}, LSCD consistently ranks among the fastest coverage criteria across nearly all benchmarks with LSC and MDSC closely following behind. Across studied datasets, LSCD is the fastest in most cases, requiring 112 minutes for MNIST, 128 minutes for GTSRB, and 126 minutes for CIFAR-100. A minor exception occurs for SVHN, where LSC is slightly faster (144 minutes) than LSCD (146 minutes), with a negligible difference of 2 minutes. MDSC also demonstrates comparable efficiency, consistently ranking close to LSC across all datasets.

The relative efficiency of these metrics can be attributed to their computational methods. LSC relies on density estimation using kernel density estimation, while MDSC incorporates covariance-aware distance computation. In contrast to other non-structural metrics, DSC is computationally more expensive due to its reliance on pairwise nearest-neighbor and distance calculations. LSCD achieves its efficiency through a simpler formulation that measures distances only between latent representations and class centers.

In summary, LSCD, LSC, and MDSC form a group of computationally efficient coverage criteria, with LSCD exhibiting a consistent but moderate runtime advantage in the evaluated settings. The efficiency gap widens when compared to the structural metrics. Among the structural coverage metrics, NC is the most efficient, requiring only 3 days, 2 hours, 21 minutes across all datasets. k-MNC is the most computationally intensive criterion, requiring over 200 hours to complete the campaigns. LSCD is 8.71 and 24.6 times faster than NC and k-MNC, respectively. We further observe that computational time does not vary significantly across seed selection strategies. Therefore, we report results for Equalized Sampling here, as it consistently achieves higher seed queue utilization across all evaluated datasets. Detailed runtime results for other strategies are provided in the Appendix ~\cite{replication-package}.

\tcbset{breakable = false, enhanced jigsaw, left=1mm, right=1mm, bottom=1mm, boxsep=1mm}
\begin{tcolorbox}[title=\textbf{Answer to RQ\textsubscript{2}: 
Efficient coverage criteria for fuzzing}]
\textit{Across all evaluated fuzzing configurations, non-structural coverage criteria are consistently more computationally efficient than structural coverage criteria. Among these, LSCD, LSC, and MDSC exhibit the lowest overhead, with LSCD showing a modest runtime advantage. It provides approximately 1.16 and 1.19 times speedup over LSC and MDSC, respectively. Compared with structural criteria, they are approximately 8.71 times faster than NC (the fastest structural metric) and 24.6 times faster than k-MNC (the most computationally intensive criterion).}
\end{tcolorbox}

\subsection{RQ\textsubscript{3} (Effectiveness in cross-coverage maximization)}
Fuzzing is fundamentally designed to maximize the coverage criteria used as the objective function. For example, when neuron coverage (NC) is used, the fuzzer primarily aims to maximize neuron activation. However, a significant characteristic often observed in DNNs is that optimizing one particular coverage criterion can increase the values of other coverage criteria as well, a phenomenon we refer to as "cross-coverage". This is due to a DNN's hierarchical structure and the complex interplay of its neurons. 

\begin{table*}[h]
  \centering
  \caption{Coverage achieved by original and corner case data sampled using various coverage criteria during fuzzing. \\ \footnotesize Note: The maximum achieved coverage is highlighted in green.} 
  \label{tab:coverage_comparison}
  \renewcommand{\arraystretch}{0.6}
  \begin{adjustbox}{width=\linewidth}
\begin{tabular}{c c c c c c c c c c c} 
      \toprule
      \multirow{2}{*}{Dataset} & \multirow{2}{*}{Type} & \multirow{2}{*}{\shortstack{Split used \\ for evaluation}} & \multicolumn{4}{c}{Structural coverage criteria} & \multicolumn{4}{c}{Non-structural Coverage criteria}\\
      \cmidrule(lr){4-7} \cmidrule(lr){8-11}
      & &  & NC & k-MNC & NBC & SNAC & LSC & DSC & MDSC & LSCD \\
    \midrule
    \multirow{8}{*}{MNIST} & Original & Test Dataset & 69.49 & 76.76 & 2.18 & 3.09 & 40.60 & 43.00 & 53.90 & 25.27 \\
    \cmidrule(lr){2-11}
    &  \multirow{7}{*}{\shortstack{Corner Case Data \\generated \\ using-}} & NC & 74.57 & 79.99 & \cellcolor{green!20} 17.97 & 25.5 & 25.7 & 61.60 & 91.60 & 57.97\\
    &   &  k-MNC & 75.84 & 80.2 & 15.4 & 26.1 & 29.60 & 60.90 & 90.20 & 58.12 \\
    &  &  NBC & 75.84 & 80.3 & 15.59 & 26.39 & 37.60 & 63.30  & 93.30 & 58.07\\
    &  &  LSC & 75.84 & 80.73 & 16.31 & 27.03 & 31.60 & 61.40 & \cellcolor{green!20} 93.90 & 58.28 \\
    &  &  DSC & 75.84 & 80.73 & 16.31 & 27.03 & 20.10 & \cellcolor{green!20} 67.90 & 93.50 & 58.40 \\
    &  &  MDSC & \cellcolor{green!20} 76.69 & 80.31 & 15.67 & 26.44 & 33.40 & 64.40 & 93.20 & 58.67 \\
    &   & LSCD & 75.84 & \cellcolor{green!20} 80.87 & 16.50 & \cellcolor{green!20} 27.33 & \cellcolor{green!20} 41.00 &  62.90 & 90.10 & \cellcolor{green!20} 58.69 \\
      \midrule
      
\multirow{8}{*}{SVHN} & Original Dataset & Original Dataset & 57.48 & 69.44 & 3.10 & 3.20 & 68.60 & 85.90 & 53.40 & 6.14 \\
    \cmidrule(lr){2-11}
    & \multirow{7}{*}{\shortstack{Corner Case Data \\generated \\ using-}} & NC & 76.48 & 82.76 & 29.01 & 22.8 & \cellcolor{green!20}56.60 & 52.90 & 52.20 & 11.14 \\
    &   &  k-MNC & 78.17 & 85.15 & 33.2 & 26.37 & 55.60 & 47.90 & 51.50 & \cellcolor{green!20} 11.16 \\
    &  &  NBC & 79.51 & 86.60 & 36.06 & 28.87 & 49.90 & 53.70 & 62.90 & 11.14 \\
    &  &  LSC & 80.22 & 87.57 & 38.00 & 30.47 & 52.50 & 58.00 & 49.70 & 11.15 \\
    &  &  DSC & 80.72 & 88.19 & 39.37 & 31.58 & 51.00 & 60.80 & 52.10 & 11.14 \\
    &  &  MDSC & 76.34 & 82.12 & 28.25 & 22.07 & 54.30 & 56.90 & \cellcolor{green!20} 65.70  & 11.15 \\
    &   & LSCD & \cellcolor{green!20} 81.14 & \cellcolor{green!20} 88.66 & \cellcolor{green!20} 40.29 & \cellcolor{green!20} 32.45 & 53.70 & \cellcolor{green!20} 60.90 & 61.70 & 11.15\\
      \midrule
    \multirow{8}{*}{GTSRB} & Original Dataset & Original Dataset & 52.17 & 75.62 & 3.68 & 4.28 & 49.10 & 84.7 & 61.30 & 55.71 \\
    \cmidrule(lr){2-11}
    & \multirow{7}{*}{\shortstack{Corner Case Data \\generated \\ using-}} & NC & 81.78 & 83.07 & 30.46 & 34.62 & 28.70 & 45.20 & 67.70 & 193.27\\
    &   &  k-MNC & 83.43 & 84.83 & 33.9 & 37.03 & 22.70 & 55.70 & \cellcolor{green!20} 83.10 & 193.42 \\
    &  &  NBC & 84.19 & 85.86 & 36.32 & 38.44 & 32.80 & 52.50 & 77.20 & 193.2 \\
    &  &  LSC & 84.81 & 86.74 & 38.22 & 39.51 & 32.50 & 55.80 & 82.20 & 192.47 \\
    &  &  DSC & 85.02 & 87.37 & 39.79 & 40.41 & \cellcolor{green!20}34.30 & 60.50 & 71.60 &192.48 \\
    &  &  MDSC & 82.33 & 83.54 & 31.54 & 35.77 & 31.60 & 47.80 & 73.20  & 193.42 \\
    &  & LSCD & \cellcolor{green!20} 85.57 & \cellcolor{green!20} 87.89 & \cellcolor{green!20} 41.18 & \cellcolor{green!20} 41.20 & 33.90 & \cellcolor{green!20} 60.70 & 78.20 & \cellcolor{green!20} 193.77 \\
  \midrule
    \multirow{8}{*}{CIFAR-100} & Original Dataset & Original Dataset & 44.89 & 61.94 & 2.01 & 1.39 & 14.00 & 15.50 & 70.70 & 17.05 \\
    \cmidrule(lr){2-11}
    & \multirow{7}{*}{\shortstack{Corner Case Data \\generated \\ using-}} & NC & 63.10 & 74.92 & 15.76 & 10.79 & \cellcolor{green!20} 41.10 & 52.20 & 83.40 & 46.11 \\
    &   &  k-MNC & 63.68 & 75.83 & 17.08 & 11.73 & 28.40 & 42.50 & 83.30 & 46.13 \\
    &  &  NBC & 65.25 & 78.12 & 21.47 & 15.84 & 29.20 & 53.70 & 76.20 & \cellcolor{green!20} 46.26 \\
    &  &  LSC & 65.60 & 78.85 & 22.71 & 16.98 & 36.50 & 52.00 & 85.80 & 46.22 \\
    &  &  DSC & 65.90 & 79.39 & 23.64 & 17.78 & 36.20 & \cellcolor{green!20} 55.00 & \cellcolor{green!20} 76.70 & 46.21 \\
    &  &  MDSC & 62.36 & 75.35 & 16.54 & 11.29 & 38.70 & 53.00 & 80.10 & 46.13 \\
    &   & LSCD & \cellcolor{green!20} 66.21 & \cellcolor{green!20} 79.83 & \cellcolor{green!20} 24.43 & \cellcolor{green!20} 18.49 & 35.50 & 53.20 & \cellcolor{green!20} 77.60 & 46.20 \\
      \bottomrule
    \end{tabular}
  \end{adjustbox}
\end{table*}

The results of our cross‑coverage analysis are summarized in \autoref{tab:coverage_comparison}. Overall, the table indicates that coverage effectiveness tends to converge across different guidance criteria. In most cases, fuzzing guided by one coverage metric achieves comparable values for other criteria, regardless of whether they are structural or non-structural. When a metric does not achieve the maximum cross-coverage value, its results typically remain within a small margin (e.g., typically within 5\%) of the best-performing configuration.

In the high-dimensional setting of CIFAR-100, similar trends are observed across different guidance criteria. While structural metrics like k-MNC reach a maximum of 75.83\%, fuzzing campaigns driven by LSCD achieve nearly 79.83\% k-MNC on its own. LSCD also scales NC simultaneously to near its peak of 66.21\%. This indicates that guiding fuzzing toward untested/sparse regions of the latent space generally leads to broad exploration of DNN behavior.

To further validate this, we monitored strong neuron activation coverage, which represents the extremely high-activation values of the neuron's activation distribution as an independent measure not used during fuzzing. This provides a reliable indicator of the effectiveness of cross-coverage maximization. Across all configurations, SNAC values also show similar increases, suggesting that different guidance criteria can trigger extreme neuron activations even without explicitly using this metric during fuzzing. 

Taken together, these findings show that cross-coverage gains are broadly similar across different criteria, while the computational costs of achieving these coverage levels vary substantially. In this context, lightweight latent-space metrics such as LSCD, LSC, and MDSC offer a favorable balance, achieving coverage outcomes comparable to those of other criteria while avoiding the overhead of explicitly monitoring multiple structural metrics.

\tcbset{breakable = false, enhanced jigsaw, left=1mm, right=1mm, bottom=1mm, boxsep=1mm}
\begin{tcolorbox}[title=\textbf{Answer to RQ\textsubscript{3}: 
Effectiveness in cross-coverage maximization}]
\textit{Our analysis shows that self‑coverage (i.e., the coverage value of the selected criterion itself) and cross‑coverage (i.e., the extent to which optimizing one criterion increases other coverage metrics) improvements are broadly comparable across all studied criteria. No single criterion consistently maximizes coverage across all metrics. However, when combined with computational efficiency, LSCD, MDSC, and LSC offer a favorable trade-off between efficiency and effectiveness.}
\end{tcolorbox}

\subsection{RQ\textsubscript{4} (Generalization \& Stability)} \label{results:rq4}

To evaluate the practical utility of the generated corner cases, we performed a retraining study to measure the trade-off between generalization robustness and model stability. For each coverage criterion, we retrained the DNN using only the corner cases it produced and measured generalization robustness on a held‑out fuzzing split. As the original test set is not used during retraining, it serves as an independent benchmark for evaluating DNN's stability on the original data distribution. 

\begin{figure}[h]
  \centering
  \begin{subfigure}[b]{0.24\linewidth}
    \centering
    \includegraphics[width=\linewidth, height=1.5cm, keepaspectratio]{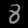}
    \caption{MNIST}
    \label{fig:mnist_cc_img}
  \end{subfigure}
  \hfill
  \begin{subfigure}[b]{0.24\linewidth}
    \centering
    \includegraphics[width=\linewidth, height=1.5cm, keepaspectratio]{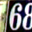}
    \caption{SVHN}
    \label{fig:svhn_cc_img}
  \end{subfigure}
  \hfill
  \begin{subfigure}[b]{0.24\linewidth}
    \centering
    \includegraphics[width=\linewidth, height=1.5cm, keepaspectratio]{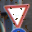}
    \caption{GTSRB}
    \label{fig:gtsrb_cc_img}
  \end{subfigure}
  \hfill
  \begin{subfigure}[b]{0.24\linewidth}
    \centering
    \includegraphics[width=\linewidth, height=1.5cm, keepaspectratio]{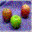}
    \caption{CIFAR-100}
    \label{fig:cifar100_cc_img}
  \end{subfigure}
  \caption{Corner-case examples generated across the four studied benchmarks.}
  \label{fig:corner_cases_all}
\end{figure}

Some samples from the generated corner cases are illustrated in \autoref{fig:corner_cases_all}. For example, the original SVHN digit "6", which remains clearly identifiable to a human, is misclassified as "5" after the fuzzer applies an autocontrast perturbation \autoref{fig:svhn_cc_img}. Similar misclassifications across other benchmarks, as illustrated in \autoref{fig:corner_cases_all}, show that the fuzzing campaign can generate challenging corner cases that expose vulnerabilities of the DNN under test. 

\begin{figure}[h]
  \centering
  \begin{subfigure}[b]{0.48\linewidth}
    \centering
    \includegraphics[width=\linewidth, keepaspectratio]{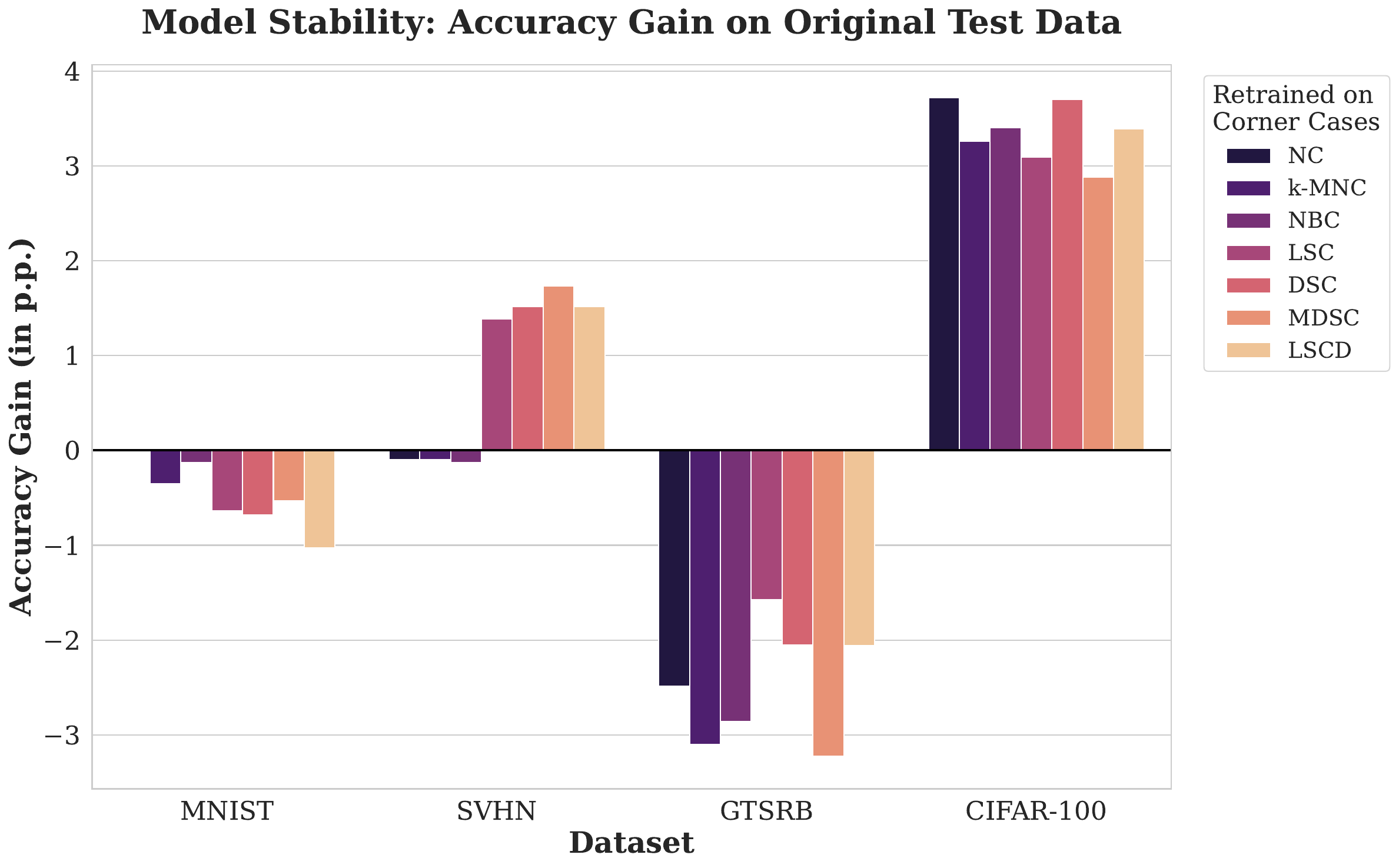}
    \caption{DNN's stability gain}
    \label{fig:dnn_stability}
  \end{subfigure}
  \hfill
  \begin{subfigure}[b]{0.48\linewidth}
    \centering
    \includegraphics[width=\linewidth, keepaspectratio]{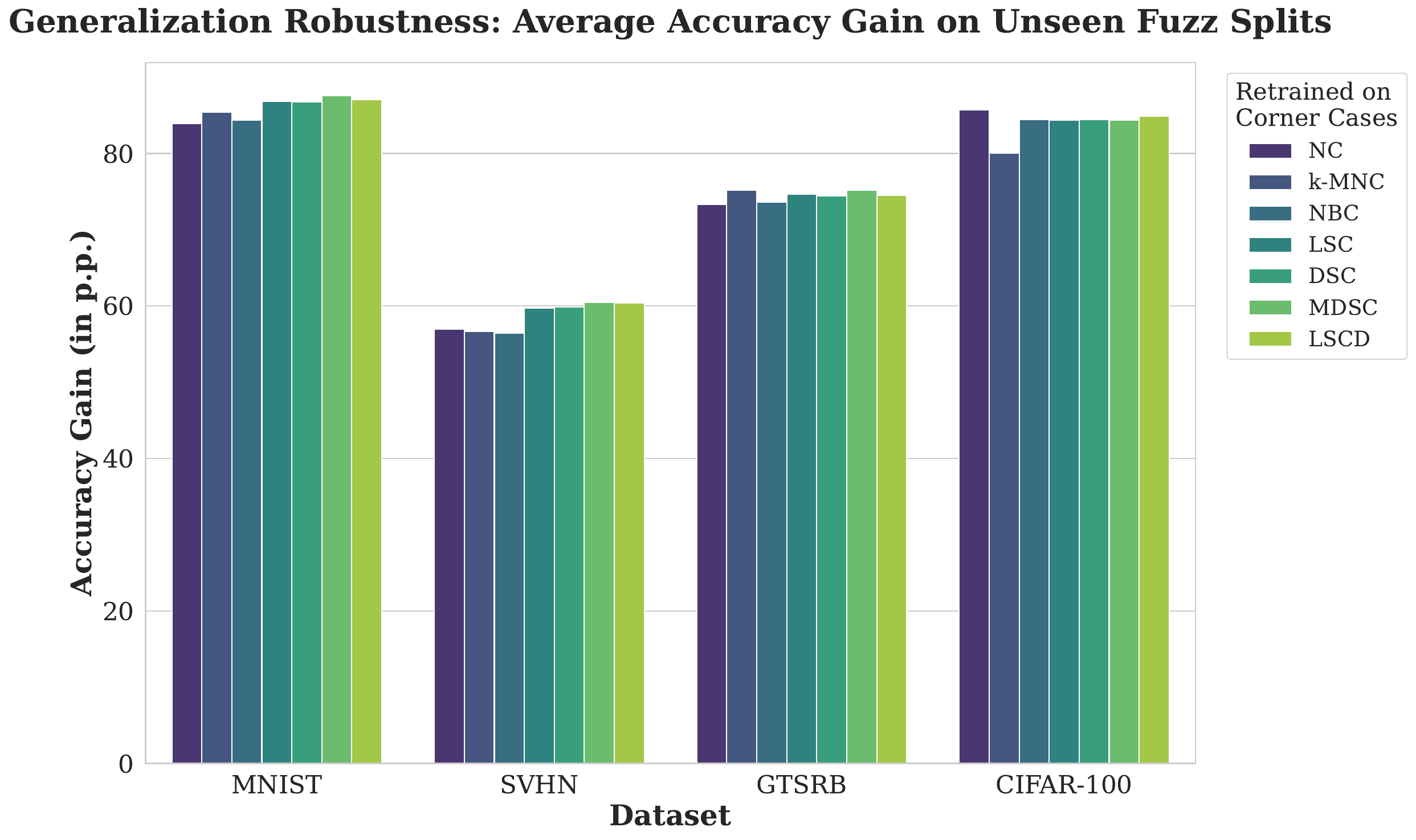}
    \caption{Generalization robustness gain}
    \label{fig:dnn_generalization}
  \end{subfigure}
  \caption{Impact of retraining using corner case data generated using CGF on DNN performance.}
  \label{fig:robustness_tradeoff}
\end{figure}

To summarize the overall DNN hardening effect, \autoref{tbl:dnn_robustness_table} reports the average accuracy across all evaluated coverage criteria. The corresponding per-criterion robustness and stability trends are shown in \autoref{fig:dnn_stability} and \autoref{fig:dnn_generalization}. \autoref{tbl:dnn_robustness_table} shows that the baseline DNNs exhibit poor performance, with a mean accuracy of 2.82\% when evaluated on the generated corner cases. This significant drop in accuracy highlights the critical need for the model-hardening strategies.

\begin{table}[H]
\centering
\caption{RQ$_4$: Impact of Retraining using Corner Cases on DNN's Robustness \\ \footnotesize Note: Baseline and retrained values report mean accuracy evaluated on corner cases (in \%), and gains report absolute accuracy differences in percentage points. Results are averaged across all evaluated coverage criteria because per-criterion differences are small.} 
\label{tbl:dnn_robustness_table}
\begin{tabular}{lcccc}
\toprule
\textbf{Dataset} & \textbf{Corner Cases} & \multicolumn{2}{c}{\textbf{Mean Accuracy (in \%)}} & \textbf{Gain (in p.p.)} \\
\cmidrule{3-4}
& & \textbf{Baseline} & \textbf{Retrained} & \\
\midrule
MNIST & 163,327 & 2.66 & 89.44 & +86.78 \\
SVHN & 261,509 & 7.12 & 68.63 & +61.51 \\
GTSRB & 190,984 & 1.13  & 77.94 & +76.81\\
CIFAR-100 & 923,289 &  0.39 & 85.44 & +85.05\\
\midrule
\textbf{Average} & \textbf{1,539,109} & \textbf{2.82} & \textbf{80.36} & \textbf{+77.53}\\
\bottomrule
\end{tabular}%
\end{table}

Retraining using these fuzzed corner cases yields substantial improvements in model robustness across all benchmarks. As shown in \autoref{fig:dnn_generalization}, retraining with fuzz-generated corner cases leads to substantial robustness gains across all evaluated coverage criteria, with only modest differences among criteria. \autoref{tbl:dnn_robustness_table} highlights accuracy gains of 86.78 p.p. on MNIST and 85.05 p.p. on CIFAR-100, while SVHN and GTSRB show improvements of approximately 61 p.p. and 76 p.p., respectively. These substantial gains are attributed to the low initial accuracy of the baseline DNNs on these corner cases, which ranged from 0\% to 7\%. The performance delta between various coverage criteria is negligible when calculating accuracy gains. This uniformity suggests that once a diverse set of corner cases is generated, the specific coverage metric used as an objective function has limited impact on the final robustness peak.
A critical concern in model hardening is the "stability margin"- ensuring that accuracy on the original, nominal test data does not degrade significantly. \autoref{fig:dnn_stability} illustrates that our retraining approach maintains high stability, with most fluctuations remaining within a negligible $\pm$3 p.p. range. Notably, for CIFAR-100, retraining on corner cases actually improved original test performance by over 3.6 p.p. This suggests that the fuzzed data helped the VGG16 model generalize better to the base distribution. The largest performance "drop" was observed in GTSRB, where accuracy on original data decreased by approximately 2.01 p.p. to 3.1 p.p. depending on the criterion used. This reduction is likely due to the model’s extremely high baseline accuracy of nearly 99.69\%, indicating that the model was already close to overfitting. Therefore, it is more sensitive to parameter updates during retraining. In MNIST and SVHN, the impact was near zero, with regressions remaining below 1 p.p.

\tcbset{breakable = false, enhanced jigsaw, left=1mm, right=1mm, bottom=1mm, boxsep=1mm}
\begin{tcolorbox}[title=\textbf{Answer to RQ\textsubscript{4}: 
Generalization \& Stability}]
\textit{Retraining with fuzzing-generated corner cases improves generalization robustness substantially, with accuracy gains of up to 86.78 p.p. across the evaluated benchmarks, while preserving DNN performance on the original test distribution with near‑zero stability change.}
\end{tcolorbox}

\subsection{Ablation Analysis: Isolating the Effects of Seed Selection and Coverage Guidance on DNN Hardening} \label{results:ablation-study}

\begin{figure}[htbp!]
  \centering
  \begin{subfigure}[b]{0.48\linewidth}
    \centering
    \includegraphics[width=\linewidth, keepaspectratio]{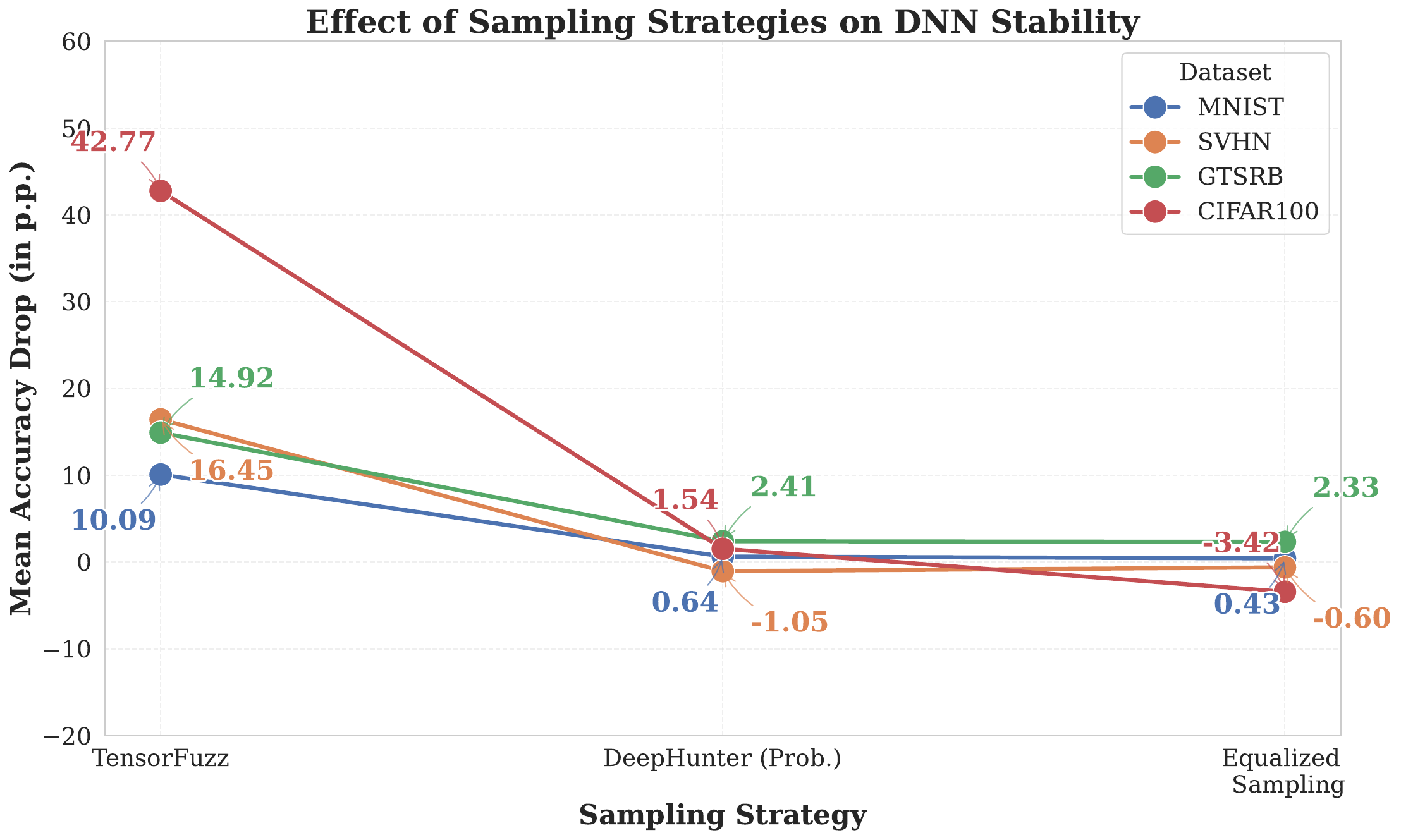}
    \caption{DNN's stability gain}
    \label{fig:ablation_study_stability}
  \end{subfigure}
  \hfill
  \begin{subfigure}[b]{0.48\linewidth}
    \centering
    \includegraphics[width=\linewidth, keepaspectratio]{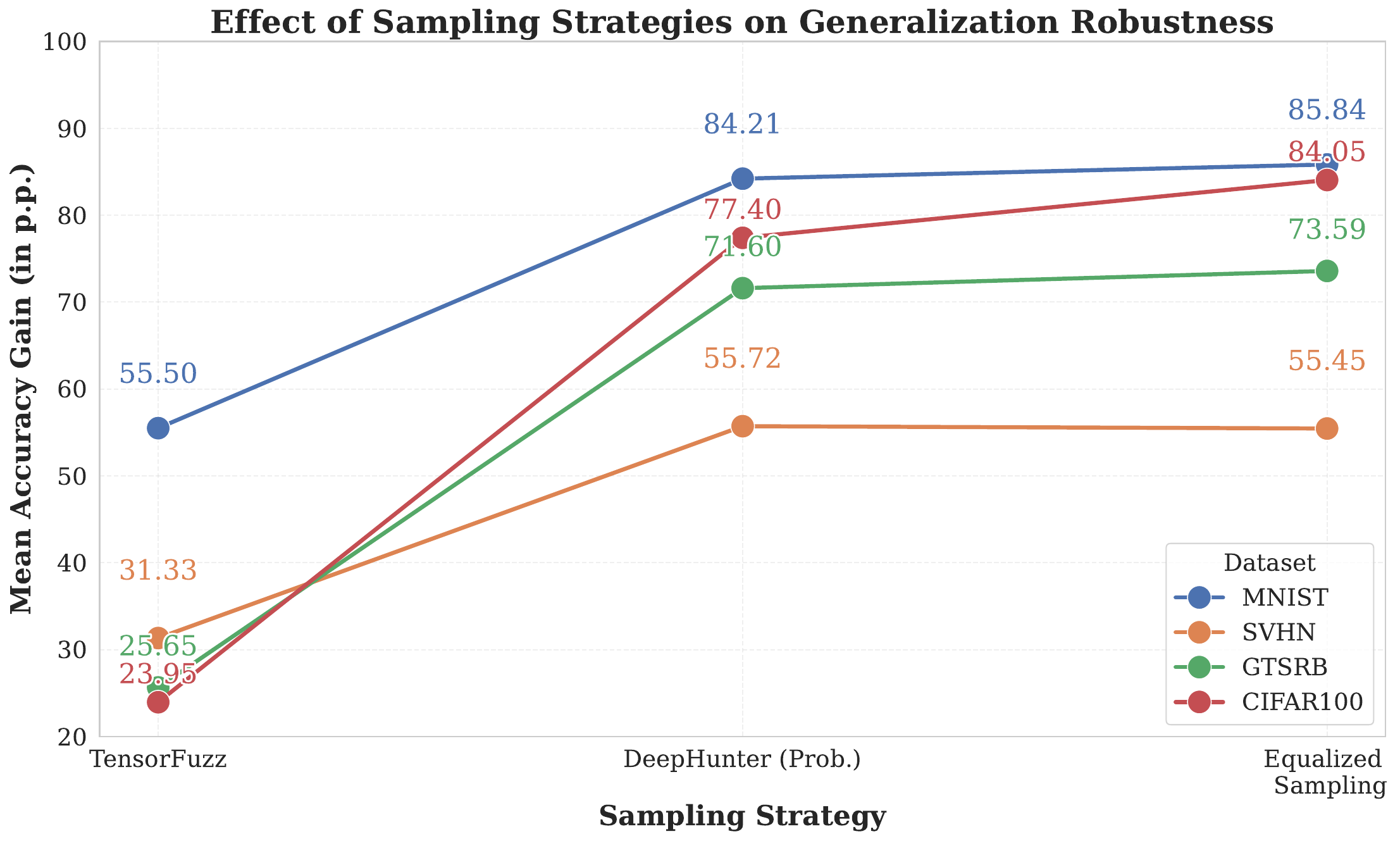}
    \caption{Generalization robustness gain}
    \label{fig:ablation_study_generalization}
  \end{subfigure}
  \caption{Impact of seed selection strategies on robustness and stability of retrained DNNs.}
  \label{fig:ablation_study}
\end{figure}

To further disentangle the impact of fuzzing configuration choices on model hardening, we conduct an ablation study separating the effects of coverage criteria and seed selection strategies. ~\autoref{fig:ablation_study_stability} and ~\autoref{fig:ablation_study_generalization} report DNN stability and mean robustness gains computed over all evaluated coverage metrics for each seed selection strategy. As shown in \autoref{fig:ablation_study_generalization}, all seed selection strategies lead to substantial improvements in robustness through retraining. TensorFuzz consistently yields the lowest robustness gains, which is consistent with its low seed queue utilization observed in RQ\textsubscript{1}. TensorFuzz tends to repeatedly mutate a relatively small subset of parent seeds. This may limit the diversity of generated corner cases and, in turn, provide less diverse data for retraining. DeepHunter's probability-based selection and Equalized Sampling achieve more comparable robustness gains across most datasets. On less complex datasets such as MNIST and GTSRB, robustness gains achieved by them remain closely clustered, with Equalized Sampling exhibiting only marginal improvements. DeepHunter attains slightly higher robustness gains than Equalized Sampling in SVHN, suggesting that probability‑based prioritization may be effective when the input space is relatively constrained. However, the results on CIFAR‑100 reveal a different trend. For CIFAR-100, Equalized Sampling achieves an average robustness gain of 84.05 p.p., representing a 6.65 p.p. and 60.10 p.p. improvement over the DeepHunter and TensorFuzz baselines, respectively. This performance gap suggests that as intra‑class variance and input dimensionality increase, the ability to draw corner cases from a diverse set of parent seeds becomes more important. 

Additional benefits of using diverse seeds during fuzzing are reflected in the preservation of the stability margin, as shown in \autoref{fig:ablation_study_stability}. The TensorFuzz baseline exhibits substantial accuracy degradation on the original distribution, averaging a 21.05 p.p. drop and reaching 42.77 p.p. on CIFAR‑100, indicating that repeatedly mutating a narrow subset of seeds can lead to overfitting during retraining. In contrast, both DeepHunter and Equalized Sampling maintain stable performance on the original test distribution across most datasets. While DeepHunter exhibits stability comparable to Equalized Sampling on MNIST, GTSRB, and SVHN, Equalized Sampling shows clearer stability advantages on more complex datasets, most notably CIFAR‑100. In this setting, Equalized Sampling achieves a near‑zero average stability change of 0.31 p.p. and even improves baseline accuracy by 3.42 p.p., indicating that the retrained models generalize better to the nominal data distribution.

These results show that generating corner cases from a diverse set of parent seeds during fuzzing leads to effective corrective training data that improves DNN robustness without harming stability. We found that the choice of coverage criterion alone has little impact on this outcome; coverage is effective only when combined with an appropriate seed selection strategy. As a result, practical DNN hardening benefits most from pairing computationally efficient coverage metrics with diversity‑aware seed sampling.

\subsection{Actionable Guidelines for DNN Fuzzing}
Based on our empirical evaluation, we summarize the key takeaways for designing efficient and effective DNN testing pipelines:

\begin{itemize} 

\item \textbf{Prefer Computationally Efficient Coverage Criteria When Effectiveness Is Comparable:} Practitioners operating under compute or model-access constraints should consider lightweight non-structural metrics such as \textit{LSCD, LSC, and MDSC} as fuzzing guidance criteria. Our results show that coverage and robustness outcomes are broadly comparable across several criteria, whereas computational costs differ substantially. This makes non-structural metrics a practical choice for scalable fuzzing, particularly in gray-box settings where full access to internal neuron activations may be limited.

\item \textbf{Prioritize Input Diversity :} 
Our results indicate that seed selection strategies influence how effectively fuzzing leverages the available input seeds. In particular, seed queue utilization provides a useful diagnostic for identifying whether the mutation budget is concentrated on a small subset of seeds or distributed more broadly across the available seed pool. Equalized Sampling offers a simple, constraint-based mechanism to improve utilization by reducing disproportionate focus on frequently selected seeds. While its gains over stronger baselines such as DeepHunter are moderate, it helps ensure that a larger fraction of available seeds receives mutation budget and can contribute to corner-case discovery.

\item \textbf{Monitor The Balance Robustness and Stability:} Fuzz-generated corner cases are useful for DNN hardening if they improve robustness without substantially degrading performance on nominal inputs. Practitioners should therefore evaluate retrained models on both held-out fuzzed data and the original test distribution. In our experiments, retraining with fuzz-generated corner cases improves generalization robustness while maintaining near-zero changes in original test accuracy across most configurations. This suggests that fuzz-generated data can serve as corrective training data, provided that stability on the original distribution is explicitly monitored.
\end{itemize}

\subsection{Threats to Validity} 

\subsubsection{Internal validity} 
We carefully selected NC, k-MNC, and NBC coverage criteria from the spectrum of structural coverage criteria for fuzzing in our empirical study. This is because other structural metrics, such as SNAC, are fundamentally subsets or specialized variations of the selected criteria. However, we used SNAC as an independent "unseen" baseline to evaluate the effectiveness of cross-coverage maximization. Another potential threat is the absolute computational time reported in this study. Although it is a function of our specific hardware configuration, it does not impact the relative efficiency rankings of the criteria. As the total fuzzing time can be optimized through more advanced hardware or distributed computing, the intrinsic design and algorithmic complexity of the coverage criteria remain constant. Consequently, while the absolute duration might decrease with superior resources, the computational overhead will scale proportionally across all criteria, preserving the validity of our conclusions.

\subsubsection{External validity}
In our study, we used well-known and widely studied DNN models such as LeNet, Spatial Transformer Networks, and VGG-16 ~\cite{lenet_5, gtsrb_repo, vgg16}. We also carefully selected popular datasets based on their varied characteristics and applicability to real-world scenarios, particularly automated driving. Despite their popularity, CIFAR-10 has been criticized for its low resolution and ambiguous labeling, which can hinder both their use in fuzzing and human recognizability of sampled corner cases ~\cite{cifar_10_quality, cifar10_1}. Therefore, we focused on working with other datasets consisting of high-quality images. 

To ensure our findings generalize to complex, high-dimensional tasks, we integrated CIFAR-100 in our experimental suite. CIFAR-100 represents a significantly more challenging domain, characterized by high inter-class variability and a broad input space. While many studies utilize only a small subset of the 1,000 ImageNet classes to mitigate computational costs, our deployment of the full 100-class distribution in CIFAR-100 serves as a high-fidelity proxy for large-scale classification challenges. This rigorous setup allows us to evaluate how various coverage criteria scale under the high inter-class variance typical of real-world perception systems. 

\subsubsection{Transferability beyond Computer Vision}
Our experiments focus on computer vision, but the methodology and results target properties that are not inherently tied to images and are transferable to other domains, such as Natural Language Processing (NLP). Structural coverage metrics (e.g., NC, $k$-MNC) depend on DNN architecture. Therefore, their role as test adequacy proxies is conceptually similar across tasks. Similarly, non-structural metrics operate on latent space distributions, making their logic agnostic to whether the features are derived from images or text. However, we acknowledge that 'mutations' in NLP (e.g., word substitutions) differ from pixel-level perturbations. We intend to explore these discrete, sequential domains in future work to validate our findings.

\subsubsection{Reproducibility}
Our complete fuzzing framework, along with generated corner case data, is available in our replication package ~\cite{replication-package}. This package also contains sampled corner case data with associated metadata and a validity assessment report, significantly enhancing its utility for future reuse by other researchers.

\section{Related Work} \label{Related Work}
As highlighted in various industry surveys, collecting and labeling a test dataset is a cost and resource-intensive effort~\cite{ml_testing_survey, structured_verification_ml_rahul, 2020-Riccio-EMSE}. At the same time, a diverse test dataset that broadly captures real-world operating conditions is necessary to assess the reliable performance of DNNs in real-world applications. To address this challenge, automated test generation methods have been proposed to identify corner cases that reveal defects in the DNN under test. In particular, coverage-guided fuzzing has emerged as a promising approach for generating challenging test inputs from existing datasets. Several prominent fuzzing frameworks, such as DeepHunter~\cite{deephunter}, DeepTest~\cite{deeptest}, DeepXplore~\cite{deepxplore}, and DeepGauge~\cite{deepgauge}, have contributed to DNN test generation and are referenced in industrial testing practice~\cite{testing_ml_industry_icse21}. However, despite these advances, there is still limited guidance on selecting efficient and effective fuzzing configurations in practice. This includes choosing appropriate coverage criteria and seed selection strategies, both of which can significantly influence the fuzzing outcome. Therefore, our empirical study systematically analyzes fuzzing configurations across seed selection strategies and structural/non-structural coverage criteria to derive practical guidance for scalable and effective DNN fuzzing.

We first identify a critical but often overlooked bottleneck in fuzzing effectiveness: seed queue utilization. Recent advancements, such as DeepGini  ~\cite{deepgini}, and automated prioritization frameworks ~\cite{deephyperion, seed_selection, test_selection_methods}, have significantly improved the ability to provide high-quality initial test seeds from available datasets. However, even when provided with such high-quality inputs, the final effectiveness of a fuzzing campaign is ultimately dictated by the fuzzer’s internal selection strategy, which determines which seeds are actually prioritized for mutation. In current state-of-the-art frameworks like DeepHunter ~\cite{deephunter}, internal probability-based selection has proven more effective at identifying diverse failure modes than traditional random selection or other internal variants. DeepHunter also explores other alternatives, such as random or uniform selection. However, these variants were shown to be less effective at driving diverse coverage growth, and failure discovery ~\cite{deephunter}. Therefore, we adopt the probability-based approach as a baseline along with TensorFuzz ~\cite{tensorfuzz}. Our research uniquely addresses the seed queue utilization by employing the entire available test dataset rather than a randomly sampled subset. This approach could allow us to establish a performance "ceiling," representing the maximum potential discovery rate, and identify exactly where current selection dynamics fail to leverage the full available input seeds. We also propose a constraint-based seed selection strategy to maximize seed queue utilization during fuzzing.

Coverage criteria are commonly used in DNN testing both to evaluate test dataset quality and to guide test input generation during fuzzing ~\cite{surprise_adquacy, deepxplore, deepgauge, deephunter, deeptest, lscd_intro}. Popular DNN fuzzing frameworks have primarily relied on structural coverage criteria based on various forms of neuron activations ~\cite{deepxplore, deepgauge, deephunter}. Structural coverage criteria require complete access to the DNN under test to calculate the coverage, making them difficult to apply in practice. Also, the efficacy of such structural criteria for assessing test data quality has been questioned~\cite{nc_not_meaningful, mutation_test_gen_stvr}. While Neural Coverage (NLC) was proposed as a distribution-aware criterion~\cite{nlc_paper}, recent critical reviews have shown that it may lack desirable properties such as monotonicity and order independence, and that coverage values can fluctuate substantially due to data ordering effects~\cite{nlc_revisit}. These limitations motivate the study of alternative criteria that are more scalable, interpretable, and suitable for practical fuzzing pipelines.

Prior work has shown that the latent space of a DNN encodes meaningful model behaviors, such as decision-making and memorization ~\cite{deepmemory, luna, pollmgraph}. These insights have primarily been used for DNN analysis, interpretation, and test dataset quality assessment rather than as guidance signals for test input generation. Non-structural test adequacy criteria based on latent-space behavior, such as Surprise Coverage and LSCD, characterize how test inputs relate to the training distribution in the latent space of a trained DNN ~\cite{surprise_adquacy, surprise_adequacy_journal, lscd_intro}. Recent work further shows that such non-structural metrics can approximate Mutation Score (MS) based assessment of test data quality, suggesting that latent-space behavior can provide useful insights into the defect-revealing potential of a test dataset without requiring mutant generation~\cite{lscd_intro}. Despite their popularity in test dataset quality assessment, the role of these non-structural criteria as coverage criteria within fuzzing frameworks remains largely unexplored. Our study addresses this gap by evaluating structural and non-structural criteria as guidance signals during fuzzing and analyzing their efficiency, cross-coverage behavior, and downstream utility for DNN hardening.

As mentioned, Mutation-based techniques provide another important perspective for assessing DNN test quality~\cite{deep_mutation, deep_mutation_plus, deep_crime}. Mutation score evaluates a test dataset based on its ability to expose injected faults in mutated DNNs and can therefore provide a more defect-oriented signal than coverage alone. Recent work has also explored mutation-score-guided generation of DNN test cases~\cite{mutation_test_gen_stvr}. In particular, DL-MSGen directly uses MS combined with pixel-statistics-based perturbation as a fitness objective to generate a test dataset ~\cite{mutation_test_gen_stvr}. While this approach improves mutation score and boosts adversarial robustness, calculating MS itself remains computationally expensive due to the need to generate and evaluate numerous mutant models~\cite{deep_crime, lscd_intro}. Recent studies indicate that latent space-based metrics can approximate MS without requiring mutant generation~\cite{lscd_intro}. Our work complements these insights: instead of directly optimizing MS for dataset generation, we investigate how lightweight latent-space coverage metrics can offer a more practical, resource-aware trade-off between efficiency and effectiveness for test dataset generation.

Since our focus is on coverage-guided fuzzing rather than direct mutation-score optimization, we discuss our fuzzing setup with respect to existing DNN fuzzing frameworks. DeepTest~\cite{deeptest} is an application-specific fuzzing framework that employs Neuron Coverage as a coverage criterion. DeepHunter is a more general-purpose fuzzing framework that uses metamorphic relations based on affine and pixel-value transformations to transform input images~\cite{deephunter}. In our empirical study, we focus on a diverse set of domain-relevant transformations that simulate environmental stressors commonly studied in computer vision robustness testing ~\cite{hendrycks_framework, image_perturbations_stocco}. Perturbations such as spatter noise, fog, brightness shifts, and blur are particularly relevant for safety-relevant perception tasks, e.g., Automated Driving. Prior fuzzing frameworks often rely on $L_p$-norm constraints to filter transformed inputs and avoid excessive perturbations. While such constraints are useful for controlling perturbation magnitude, they are also closely aligned with adversarial-style input generation and may restrict broader domain-relevant corruptions such as illumination changes, weather effects, or sensor noise. In contrast, our study uses domain-relevant transformations that affect the image more globally and are commonly used in robustness evaluation ~\cite{hendrycks_framework, image_perturbations_stocco}. This allows us to study fuzzing under perturbations that are more closely aligned with operational environmental variations than pixel-level adversarial changes.

While retraining on augmented or adversarial data is a standard practice in machine learning, prior work on DNN fuzzing has primarily focused on corner‑case discovery ~\cite{deephunter, deeptest, tensorfuzz}. They aimed to maximize coverage guidance to expose failure modes, but they typically treat discovered corner cases as end products and provide limited evidence of their utility for DNN hardening. Motivated by this observation, we use retraining using fuzzing-generated data as a validation step rather than a contribution. In this work, we study how fuzzing configuration choices (coverage guidance and seed selection strategies) affect fuzzing efficiency and effectiveness, and how the generated corner case data impacts DNN hardening (robustness on held-out fuzzed splits) while monitoring nominal performance stability. Overall, our study goes beyond corner‑case discovery by providing practitioners with evidence‑based configuration guidance under model‑access and compute constraints.

\section{Conclusion and Future Work} \label{Conclusion_future_work}
Fuzzing has emerged as an effective approach for generating a challenging test dataset that exposes various defects in the DNN under test. However, identifying an appropriate fuzzing configuration given the wide range of coverage criteria and seed selection strategies, under practical computing and access constraints, is challenging. To address this gap, we conducted a large‑scale empirical study of 84 different fuzzing configurations across four widely used computer vision benchmark datasets. We scrutinize various key factors that can ensure effective and efficient fuzzing. 

Our analysis shows that fuzzing effectiveness is governed by an interplay between coverage criterion and seed selection rather than by either component in isolation. At the end of the fuzzing campaign, the achieved coverage values and robustness gains became broadly comparable across the evaluated criteria. In contrast, computational costs and the degree to which the available seed pool is utilized during fuzzing vary substantially across configurations, making efficiency and input diversity critical practical differentiators.

Among the evaluated coverage criteria, LSCD, MDSC, and LSC achieve cross-coverage and model hardening outcomes comparable to those of other metrics at a fraction of the computational cost. Our proposed "Equalized Sampling" improves seed queue utilization by enabling a larger fraction of available inputs to actively participate in fuzzing. Equalized Sampling uses the maximum number of available initial seeds during fuzzing, achieving an average 77.58\% seed queue utilization across all considered datasets, resulting in 12.97 p.p. and 70.59 p.p. improvements over DeepHunter and Tensorfuzz, respectively. Together, these findings suggest that pairing lightweight coverage guidance with diversity‑aware seed selection yields a practical and resource‑aware approach to DNN fuzzing.

Finally, integrating fuzzed corner cases into retraining enhances a DNN's general robustness while preserving its stability under the original data distribution. Our retraining experiments demonstrate that DNNs hardened with corner cases achieve a significant average robustness Gain of 77.53 p.p. on unseen fuzzed test splits. Simultaneously, the retrained DNNs maintain the stability margin below 3 p.p., ensuring that performance on nominal data is not compromised.

In future work, we aim to extend this study in three directions. First, we plan to develop methods for quantifying and categorizing the faults exposed by fuzz-generated corner cases. Unlike traditional software, where a taxonomy of bugs (e.g., memory leaks, race conditions) exists, DNNs present a unique challenge as the faults may be unique to the DNN under test. Secondly, we will investigate how the generated corner cases discovered through fuzzing can be leveraged to effectively repair faults. Finally, we intend to extend fuzzing techniques beyond image classification to other DL domains such as semantic segmentation and 3D object detection. As the latent space vectors for these DNNs are more complex, we must first develop advanced methods for computing non-structural coverage criteria and then integrate them effectively into the fuzzing process.

\bibliographystyle{ACM-Reference-Format}
\bibliography{parts/9_bibliography}

\end{document}


\title{Appendix: FuzzInsight: An Empirical Study on \\Efficient and Effective Coverage-Guided Fuzzing for DNNs}

\author{Vivek Vekariya}
\email{vivek.vekariya@tum.de}
\orcid{0009-0005-1482-7142}
\affiliation{%
  \institution{Technical University of Munich}
  \city{Munich}
  \state{Bayern}
  \country{Germany}
}
\author{Fathiyeh Faghih}
\email{fathiyeh.faghih@bcu.ac.uk}
\orcid{0000-0002-8877-6895}
\affiliation{%
  \institution{Birmingham City University}
  \city{Birmingham}
  \state{West Midlands}
  \country{United Kingdom}
}
\author{Alexander Pretschner}
\email{alexander.pretschner@tum.de}
\orcid{0000-0002-5573-1201}
\affiliation{%
  \institution{Technical University of Munich \& fortiss GmbH}
  \city{Munich}
  \state{Bayern}
  \country{Germany}
}
\renewcommand{\shortauthors}{Vekariya et al.}

\maketitle

\section{Dataset and Artifacts Availability} 

The corner case data generated during our experiments is available for download at \href{https://syncandshare.lrz.de/getlink/fiPqLHNXt2Lai8zKjrPXB9/}{this link}. Due to the substantial size of the full dataset, we provide a representative sample generated using the Equalized Sampling seed-selection strategy. The remaining data can be reproduced using the fuzzing framework included in our replication package ~\cite{replication-package} or can be provided upon request. Detailed results and setup instructions are available in the repository's README file.

\section{Results} 

\autoref{tab:cc_overview} details the total number of corner cases generated by the three seed selection strategies. This breakdown across coverage criteria shows that the Equalized sampling strategy uses the most initial seeds across all datasets and criteria.

\begin{figure*}[t]
  \centering
 \begin{subfigure}[t]{0.48\textwidth}
    \centering
    \includegraphics[width=\linewidth, keepaspectratio]{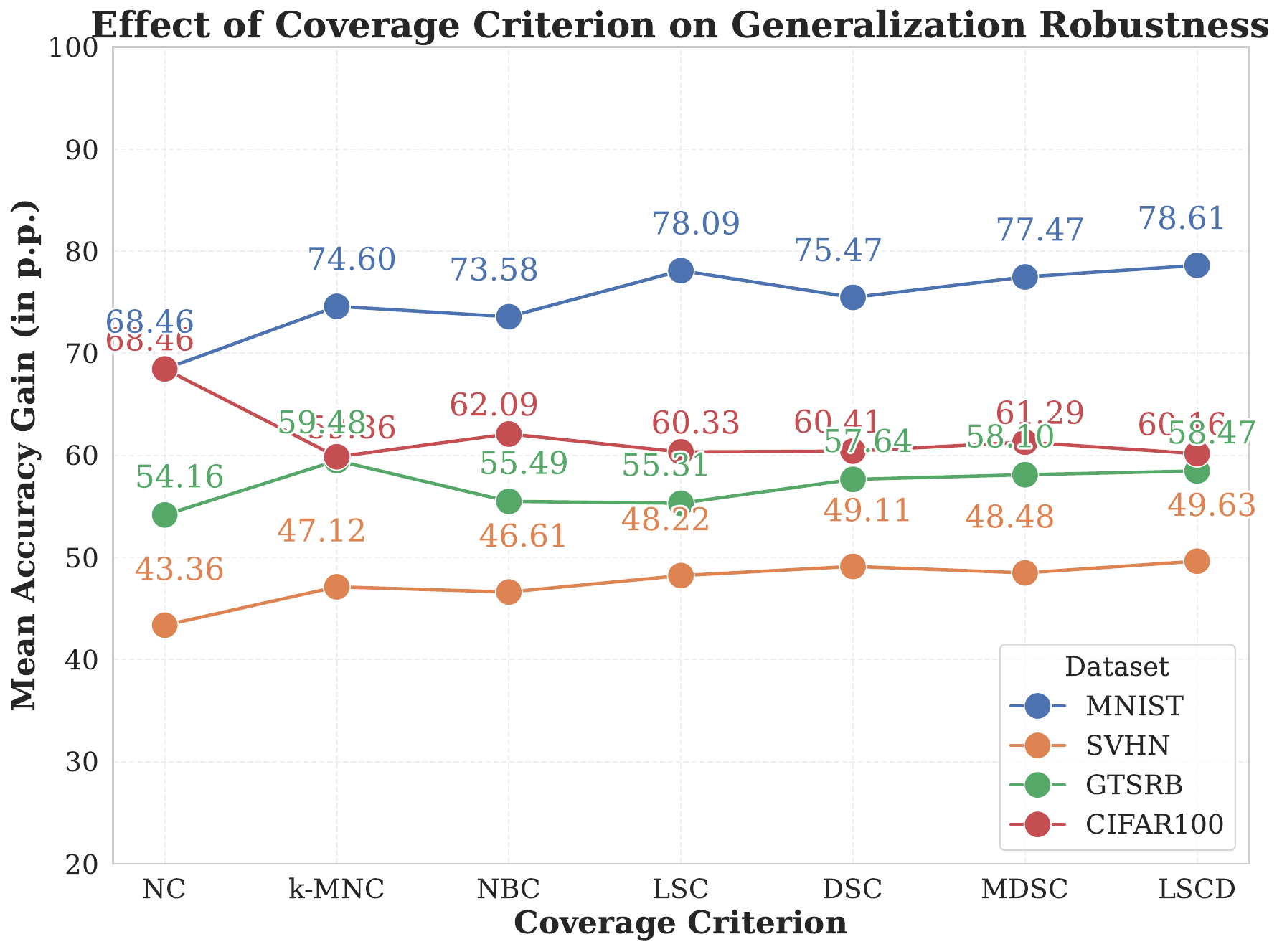}
    \caption{Generalization Robustness}
    \label{fig:per_criteria_robustness}
  \end{subfigure}
  \hfill
  \begin{subfigure}[t]{0.48\textwidth} 
    \centering
    \includegraphics[width=\linewidth, keepaspectratio]{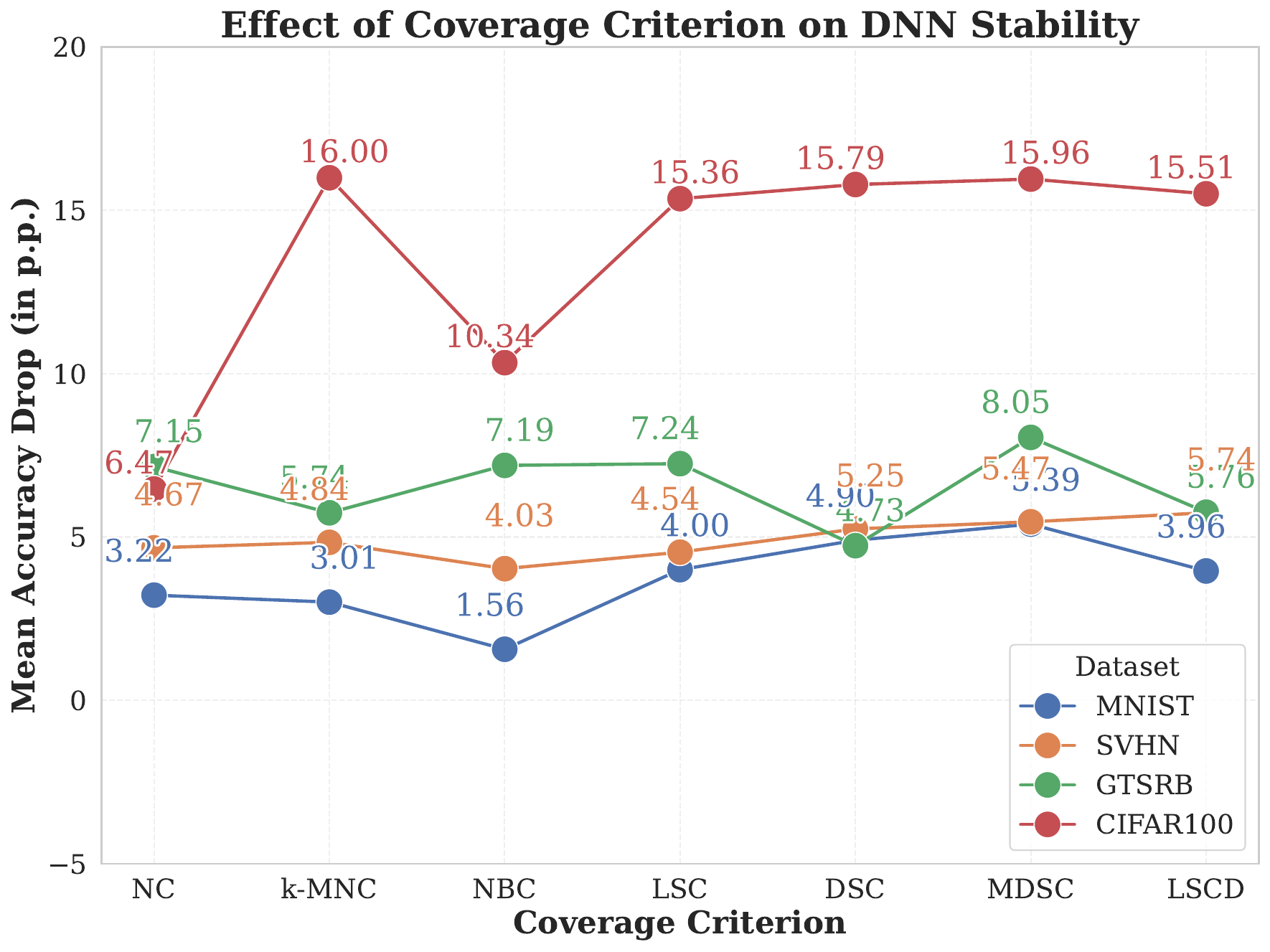}
    \caption{Stability Analysis}
    \label{fig:per_criteria_stability}
  \end{subfigure}

  \caption{Breakdown of generalization robustness and DNN stability analysis across different coverage criteria and datasets.}
  \label{fig:robustness}
\end{figure*}

Furthermore, Figures \ref{fig:per_criteria_robustness} and \ref{fig:per_criteria_stability} illustrate the accuracy gain in percentage points (p.p.), broken down by dataset and coverage criterion. As observed, the accuracy improvements/drop per criterion are marginal. A similar trend is evident regarding the gains in DNN stability

\begin{table*}[htbp]
  \centering
  \caption{Overview of identified corner case images using various seed selection strategies with the coverage criteria used during fuzzing. \\ \footnotesize Note: The cells highlighted in green depict the maximum initial seeds utilized with a specific seed selection strategy and coverage criterion.}
  \label{tab:cc_overview}
  \begin{adjustbox}{width=\linewidth}
\begin{tabular}{c c c c c c c c c }
      \toprule
      \multirow{2}{*}{Dataset} & \multirow{2}{*}{\shortstack{Coverage criteria \\used during fuzzing}} & \multirow{2}{*}{\shortstack{Total no. of \\initial seeds}} & \multicolumn{2}{c}{Tensorfuzz} & \multicolumn{2}{c}{DeepHunter - Probability} & \multicolumn{2}{c}{Equal Sampling (Our)}\\
      \cmidrule(lr){4-5} \cmidrule(lr){6-7} \cmidrule(lr){8-9}
      &  & & Corner Cases & Unique Initial Seeds & Corner Cases & Unique Initial Seeds & Corner Cases & Unique Initial Seeds \\
    \midrule
\multirow{6}{*}{MNIST} & NC & 9860 & 108785 & 1528 & 21383 & 7247 & 17741 & \cellcolor{green!20} 9014 \\
& k-MNC & 9860 & 37842 & 707 & 25379 & 5686 & 18613 & \cellcolor{green!20} 9106 \\
& NBC & 9860 & 26294 & 1394 & 21947 & 7263 & 18125 & \cellcolor{green!20} 9082 \\
& LSC & 9860 & 86306 & 354 & 26279 & 6842 & 26280 & \cellcolor{green!20} 6806 \\
& DSC & 9860 & 88671 & 317 & 27717 & 5476 & 26031 & \cellcolor{green!20} 6775 \\
& MDSC & 9860 & 88348 & 320 & 31656 & 5678 & 28130 & \cellcolor{green!20} 6871 \\
& LSCD & 9860 & 90713 & 347 & 31261 & 5726 & 28407 & \cellcolor{green!20} 6904 \\
\midrule
\multirow{6}{*}{SVHN} & NC & 24166 & 79171 & 1857 & 36321 & 11604 & 39301 & \cellcolor{green!20} 12194 \\
& k-MNC & 24166 & 50465 & 785 & 36024 & 11596 & 38217 & \cellcolor{green!20} 11959 \\
& NBC & 24166 & 35633 & 1568 & 36442 & 11537 & 38962 & \cellcolor{green!20} 12117 \\
& LSC & 24166 & 47962 & 748 & 36420 & \cellcolor{green!20} 11622 & 36439 & 11569 \\
& DSC & 24166 & 62905 & 736 & 36160 & \cellcolor{green!20} 11640 & 36332 & 11611 \\
& MDSC & 24166 & 90397 & 649 & 36814 & \cellcolor{green!20} 11643 & 36291 & 11570 \\
& LSCD & 24166 & 81934 & 653 & 36678 & \cellcolor{green!20} 11674  & 35967 & 11570 \\
\midrule
\multirow{6}{*}{GTSRB} & NC & 12593 & 89379 & 1711 & 28396 & 9315 & 25545 & \cellcolor{green!20} 11233 \\
& k-MNC & 12593 & 60657 & 547 & 30267 & 7903  & 25971 & \cellcolor{green!20} 11221 \\
& NBC & 12593 & 43549 & 1285 & 30449 & 8347 & 25668 & \cellcolor{green!20} 11210 \\
& LSC & 12593 & 35727 & 571 & 26924 & 8002 & 27069 & \cellcolor{green!20} 9833 \\
& DSC & 12593 & 32613 & 642 & 27206 & 8036 & 27329 & \cellcolor{green!20} 9909 \\
& MDSC & 12593 & 97965 & 430 & 32581 & 7834 & 29899 & \cellcolor{green!20} 9944 \\
& LSCD & 12593 & 100086 & 436 & 32126 & 7853 & 29503 & \cellcolor{green!20} 9895 \\
\midrule
\multirow{6}{*}{CIFAR-100} & NC & 7400 & 173387 & 1923 & 138117 & 7247 & 137189 & \cellcolor{green!20} 7398 \\
& k-MNC & 7400 & 179740 & 470 & 143068 & 5874  & 64003 & \cellcolor{green!20} 7161 \\
& NBC & 7400 & 143115 & 1049 & 131892 & 5852 & 136260 & \cellcolor{green!20} 7379 \\
& LSC & 7400 & 178167 & 483 & 146394 & 5876 & 144831 & \cellcolor{green!20} 7377 \\
& DSC & 7400 & 181787 & 486 & 147516 & 5848 & 144556 & \cellcolor{green!20} 7394 \\
& MDSC & 7400 & 187036 & 460 & 151052 & 5849 & 148371 & \cellcolor{green!20} 7397 \\
& LSCD & 7400 & 161272 & 445 & 151369 & 5806 & 148079 & \cellcolor{green!20} 7387 \\
      \bottomrule
    \end{tabular}
  \end{adjustbox}
\end{table*}

\bibliographystyle{ACM-Reference-Format}
\bibliography{parts/9_bibliography}